\documentclass[pdflatex]{sn-jnl}

\usepackage{graphicx}%
\usepackage{multirow}%
\usepackage{amsmath,amssymb,amsfonts}%
\usepackage{amsthm}%
\usepackage{mathrsfs}%
\usepackage[title]{appendix}%
\usepackage{xcolor}%
\usepackage{textcomp}%
\usepackage{manyfoot}%
\usepackage{booktabs}%
\usepackage{algorithm}%
\usepackage{algorithmicx}%
\usepackage{algpseudocode}%
\usepackage{listings}%

\theoremstyle{thmstyleone}%
\theoremstyle{thmstyletwo}%

\theoremstyle{thmstylethree}%

\begin{document}

\title[Article Title]{Quantum-like Coherence Derived from the Interaction between Chemical Reaction and Its Environment}


\author*[1]{\fnm{Yukio Pegio} \sur{Gunji}}\email{yukio@waseda.jp}

\author[2]{\fnm{Andrew} \sur{Adamatzky}}\email{andrew.adamatzky@uwe.ac.uk }

\author[2]{\fnm{Panagiotis}\sur{Mougkogiannis}}\email{Panagiotis.Mougkogiannis@uwe.ac.uk}

\author[3]{\fnm{Andrei}\sur{Khrenikov}}\email{andrei.khrennikov@lnu.se}

\affil*[1]{\orgdiv{Intermedia Art and Science, School of Fundamental Science and Technology}, \orgname{Waseda University}, \orgaddress{\street{Ohkubo 3-4-1}, \city{Shinjuku-ku}, \postcode{169-8555}, \state{Tokyo}, \country{Japan}}}

\affil[2]{\orgdiv{Center for Unconventional Computing}, \orgname{University of the West England}, \orgaddress{\street{Stoke Gifford}, \city{Bristol}, \postcode{BS16 1QY}, \state{State}, \country{UK}}}

\affil[3]{\orgdiv{Department of Mathematics}, \orgname{Linnaeus University}, \orgaddress{\street{Universitetsplatsen-1}, \city{Växjö}, \postcode{352 52}, \country{Sweden}}}


\abstract{By uncovering the contrast between Artificial Intelligence and Natural-born Intelligence as a computational process, we define closed computing and open computing, and implement open computing within chemical reactions. This involves forming a mixture and invalidation of the computational process and the execution environment, which are logically distinct, and coalescing both to create a system that adjusts fluctuations. We model chemical reactions by considering the computation as the chemical reaction and the execution environment as the degree of aggregation of molecules that interact with the reactive environment. This results in a chemical reaction that progresses while repeatedly clustering and de-clustering, where concentration no longer holds significant meaning.
Open computing is segmented into Token computing, which focuses on the individual behavior of chemical molecules, and Type computing, which focuses on normative behavior. Ultimately, both are constructed as an interplay between the two. In this system, Token computing demonstrates self-organizing critical phenomena, while Type computing exhibits quantum logic. Through their interplay, the recruitment of fluctuations is realized, giving rise to interactions between quantum logical subspaces corresponding to quantum coherence across different Hilbert spaces. As a result, spike waves are formed, enabling signal transmission. This occurrence may be termed quantum-like coherence, implying the source of enzymes responsible for controlling spike waves and biochemical rhythms.}




\maketitle

\section{Introduction}\label{sec1}

Is it possible to truly distinguish between natural computation and artificial computation~\cite{de2007fundamentals,maclennan1999field,maclennan2004natural,dodig2011significance}? As soon as we consider any natural phenomenon to be a computation, does it not become an artificial computation? Or, rather, is it possible to conceive of a natural computation by constructing it as an artificial computation while denying it? This paper proposes such a method, expressing it in a toy model of chemical reactions as natural phenomena ~\cite{gunji2024computation, gunji2025interaction}.
However, to avoid confusion, we will call artificial computation, that is, computation that can be expressed as a Turing machine~\cite{hopcroft1984turing}, automaton~\cite{hopcroft2001introduction}, or recursive function~\cite{cutland1980computability} (which are equivalent), closed computing. They are not influenced by anything outside except the input, and external perturbations are ignored, in principle. To that extent, they are called 'closed computing' in the sense that they are cut off from the outside.

In contrast, a system that constantly maintains a relationship with the outside and performs computations while being influenced by it is called "Open computing”. If we are talking about computation that interacts with the outside, many concepts have been proposed so far. If we use multiple components to achieve parallel computing, each component can be considered to interact with the other components, that is, the outside. Computations such as membrane computing \cite{paun2002membrane}and P systems \cite{martin2003tissue}are also constructed by imitating the relationship between a computer and its outside world, as in biological tissues. However, these are all recursive functions and Closed computing. Indeed, many of these, as well as proposals to use natural objects as computational carriers, such as DNA computing \cite{adleman1998computing, paun2005dna}, are all nothing more than recursive functions and Closed computing.

Here lies the significance of our proposal. When implementing a computation that is open to the environment, the environment must be considered infinite. This is impossible. Therefore, while writing down the relationship with the environment in a finite form, we simultaneously deny (and/or invalidate) it. By invalidating the self-defined relationship with the outside world, we achieve a computation that is not closed. This proposal can be said to be the computational version of Natural-Born Intelligence proposed against artificial intelligence \cite{gunji2016quantum, Adamatzky2022, gunji5209533natural}.

When considering chemical reactions, we cannot avoid the problem of how to connect micro- and macro organisms. This is the problem of how to connect computing that handle molecules and computing that handle concentration as a variable as a differential equation, and it is also a problem that has been proposed as the relationship between the master equation and the Fokker-Planck equation~\cite{carmichael2013statistical,metzler1999deriving}, and as the Gillespie algorithm~\cite{Gillespie1977, Gillespie2007,rao2003stochastic,vestergaard2015temporal}. Here, we will formulate computing when dealing with entities such as molecules as Token computing, and virtual concepts such as concentration as Type computing, and implement Open computing as the interplay of the two. In this case, Type computing reveals quantum logic (non-distributive orthomodular lattice \cite{svozil1993randomness, gunji2016quantum, gunji2022psychological}) in the sense of lattice theory \cite{davey2002introduction}. 
This research is consistent with Khrenikov's theory that quantum-like systems can be found in classical systems in an information-theoretic sense \cite{khrennikov2010ubiquitous, khrennikov2015quantum, khrennikov2023open}.

This paper is structured as follows. In Section 2, we define closed computing and open computing and discuss the issue of closed computing in terms of the interaction between computation and the computation execution environment. In Section 3, we first model the interaction between computation and the computation execution environment in chemical reactions using only Token computing. Secondly, we implement the interaction using Type computing, and show that the underlying logic of Type computing can be approximated as quantum logic.
In Section 4, we implement Open computing through the interplay of Token computing and Type computing. We discuss self-organized criticality\cite{Bak-Tang1989, Bak-Sneppen1993} found in Token computing, and the generation of pulse waves in the interplay of Token and Type computing. We discuss how pulse waves are generated as quantum coherence by utilizing this structure. We also mention the possibility that this may be the origin of enzymes, so to speak. Section 5 is a discussion, and Section 6 is a conclusion.

\section{Open computing and closed computing}\label{sec2}

\subsection{Definition}\label{subsec2}

 First, consider closed computing. Generally, the problem consists of two qualitatively different items. A computation that can determine whether the relationship between two such items, A and B, exists or not is defined as closed. However, closed computing, as defined in this paper, is defined as a part of that. Consider a recursive function as an example. A is the previous one and B is the next one, and a successor is defined such that it relates them with +1. If A is one recursive function and B is another recursive function and they are related by ``performing the computation sequentially" then function composition can be defined. In this way, all definitions that make up a recursive function determine the relation between the two items as presence, and relate them through some kinds of relationship. Therefore, it can be considered closed.
Closed computing is defined as a computation that is closed, especially when the two items are computing and its execution environment. In other words, the relationship between computing and its execution environment is determined to be related or not (Fig. 1A).

\begin{figure}[h]
    \centering
    \includegraphics[width=1.0\linewidth]{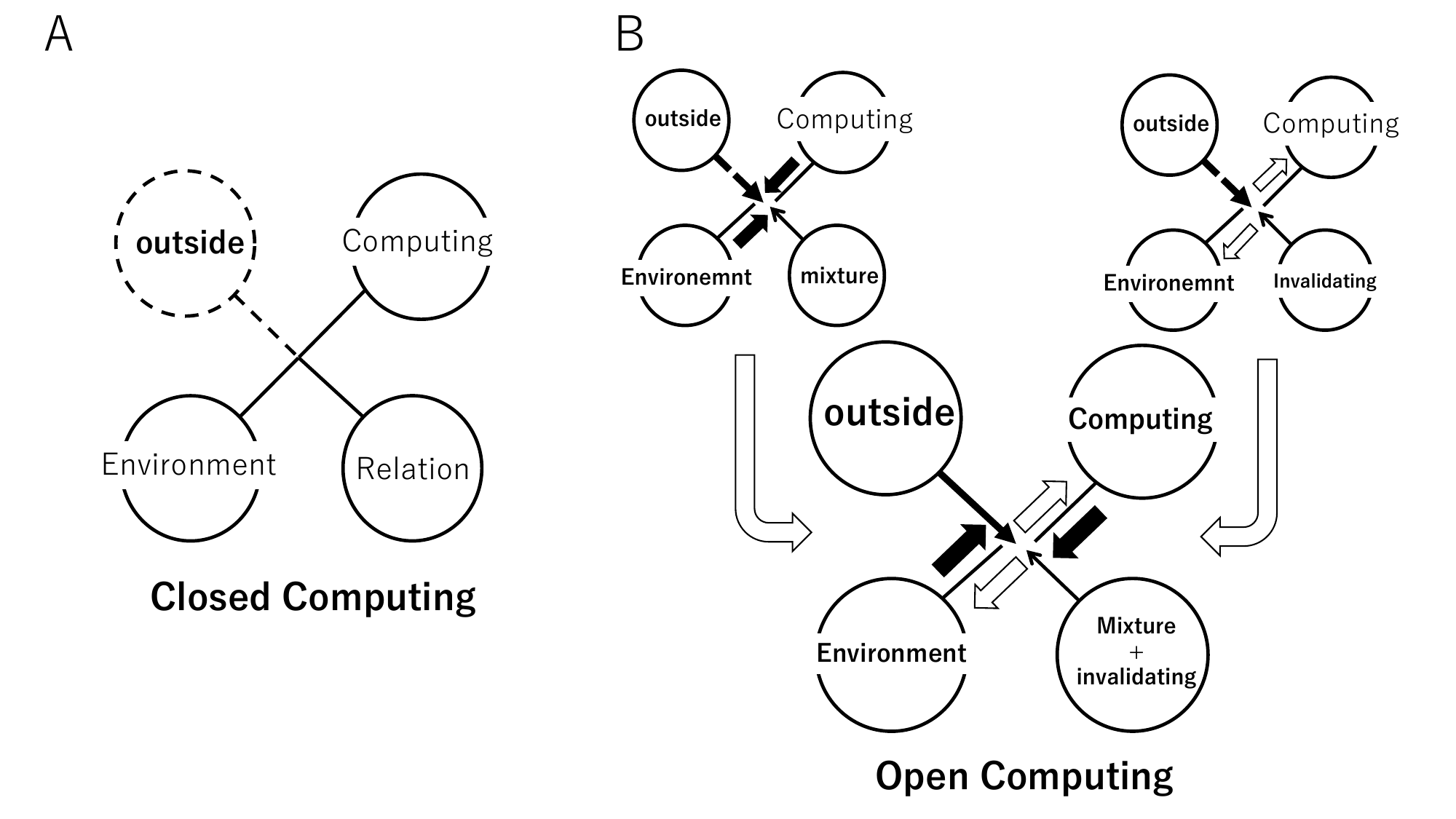}
    \caption{Schematic diagram of Closed computing and Open Computing. A. Closed computing in which computing and its environment is determined either related or not. B. Open computing consisting of mixture and invalidation of computing and its environment.}
    \label{fig:enter-label}
\end{figure}

In contrast to this, Open Computing does not make binary decisions, but rather constructs a dynamic relationship between two items by partially confusing the two and partially cancelling and invalidating the defined meanings of both. Here, the two items are computing and its execution environment. On the one hand, it confuses and connects computing and its execution environment, while on the other hand it performs an operation that invalidates the original meanings of both. Open Computing is implemented by overlapping this confusion and invalidation (Fig 1B).
By defining it in this way, Open computing maintains a dynamic state that appropriately incorporates perturbations. Especially, the Open computing of chemical reactions defined in this paper realizes a state that can be expressed by quantum logic, incorporates large fluctuations, and ultimately realizes quantum coherence.

\subsection{Uncontrollability in Closed computing}\label{subsec2}

 We first argue for the impossibility of control in closed computing. In closed computing, the relationship between computation and the computation environment is determined to exist or not, so let us assume that it does not exist. In that case, it is necessary to keep the computation environment in a healthy state for all times, regardless of the computation. Even if we limit it to the heat generated by the computer, to suppress it, the room in which the computer is placed will be constantly cooled even when the computer is stopped, which requires excessive energy, and considering this wasteful energy consumption, it is safe to say that the computation environment cannot be controlled.
  So, what if we assume that the relationship between computation and the computing environment is related? If the use of the computer causes the temperature in the room to rise, the air conditioner will detect the temperature rise and lower the temperature in the room. If such an interaction is implemented, it seems that the relationship between the computation and the computation environment can be well controlled. However, if you want to strictly manage and control the temperature, it is extremely difficult. First, the hot air generated by the computer always accumulates non-uniformly around the computer, and spatial errors become a problem when lowering the temperature in the room while taking into account the positional relationship with the air conditioner. Secondly, when trying to detect the temperature in a room and precisely control it, time delays inevitably become an issue. Considering two points, it would be extremely difficult to control the relationship between a computing device and the environment in which the computing is executed.
  The Ackermann function succinctly illustrates this difficulty \cite{kleene2007origins}. The Ackermann function was originally defined as a primitive recursive function, conceived as a function defined by recursion. It is defined as follows
  
\begin{equation}
    A(0, y)=y+1,
\end{equation}
\begin{equation}
    A(x+1, 0)=A(x, 1),
\end{equation}
\begin{equation}
    A(x+1, y+1)=A(x, A(x+1, y)).
\end{equation}

Here, the first term of the argument can be thought of as giving the state of the computation execution environment, and the second term as giving the state of the computer. In other words, the Ackermann function can be thought of as a computer that performs computation while controlling its own computation execution environment. The first equation means that when the computation execution environment reaches an appropriate state (which is represented by the value 0), the computation is completed, and a value is output. The second and third equations express the relationship between the state of the computer and the state of the computation execution environment. The states of the computer and the execution environment are defined recursively. Below is an example of such a computation execution.

\begin{eqnarray}
\begin{split}
    A(1, 2)=A(0, A(1, 1))=A(0, A(0, A(1, 0))\\
    =A(0, A(0, A(0,1)=A(0, A(0, 2))=A(0, 3)=4.
\end{split}
\end{eqnarray}

We can see that the computer can only output a value when the value of the computation execution environment becomes 0.

The Ackermann function seems to appropriately define the relationship between computation and the computation execution environment. However, if the value of the computation execution environment becomes even slightly large, the state of the computer will diverge. This can be easily seen by calculating as follows.

\begin{equation}
A(0, y)=y+1
\end{equation}
\begin{equation}
A(1, y)=y+2
\end{equation}
\begin{equation}
A(2, y)=2y+3
\end{equation}
\begin{equation}
A(3, y)+3=2^{y+3}
\end{equation}
\begin{equation}
A(4, y)+3=2^{2^{2..{^2}}}
\end{equation}

If the value of the execution environment becomes 4, nested powers appear, suggesting divergence of the computation. In fact, the Ackermann function becomes problematic due to this instability, and the primitive recursive function was redefined as a recursive function by introducing a new bounded operator to eliminate such situations. This behavior of the Ackermann function suggests that it is extremely difficult to strictly control the relationship between the computation and its execution environment.

   This problem is exactly the problem of natural computation. Living organisms must deal with the problems between the computing and the execution environment in some way while constantly performing computing in the form of chemical reactions. One way is to put a buffer between the computing and its execution environment and make the relationship between the two ambiguous. Granular computing using fuzzy control and rough sets illustrates an example of such a strategy.

   But is that enough? Making the relationship between the computing and the execution environment of the computing vague and covering it with a probability distribution means stabilizing it as a specific distribution. If the diversity of the population is suppressed by a normal distribution, outliers will be ignored. No matter how subjective the membership functions (used in fuzzy control\cite{zadeh1965fuzzy, passino1998fuzzy}) or equivalence classes (used in rough sets \cite{pawlak1982rough, pawlak2001rough}) may be, the rounding is still nothing more than an operation of stabilizing and ignoring exceptions. Do living things employ such a strategy?

   Living things sometimes behave unexpectedly for an external observer. Even if humans make a finite list of their behavior, they are always capable of betraying that assumption. It is not difficult to imagine that such things also happen in chemical reactions within living organisms. Otherwise, we would not be able to understand the behavior of insects that detect just a few pheromone molecules and fly in the direction of females\cite{butenandt1961sexuallockstoff, kaissling2001olfactory}, or the retina that senses light with a few photons\cite{hecht1942energy}. In other words, while some buffer is assumed between the computation and its execution environment, it is a dynamic buffer that is not stabilized and takes advantage of exceptional behavior in some cases. Computations that assume such dynamic buffers are what we call open computing.

\subsection{Unpacking the definition of Open computing}\label{subsec3}

The concept of Open computing described here is written in a highly abstract, almost philosophical style. In the following, we unpack the description and present it in a step-by-step manner, providing analogies and scientific parallels. In traditional computing, there are clear separations and yes/no logic: Something is either this or that, true or false, inside or outside. In Open Computing, the aim is not to maintain such sharp separations. Instead of saying “this belongs here, that belongs there,” it blurs boundaries and treats two items as interacting in a fluid way. The two entities in the quoted text are:
1. Computing – the process, the logic, the software, the algorithmic thinking.
2. Execution Environment – the hardware, the operating context, the physical or virtual substrate.
Traditionally, these are considered as separate layers. Open Computing challenges this separation, and consists of partially confusing and partially canceling:

• Partially confusing means mixing their boundaries so that we cannot cleanly say which effects come from “computation” and which from the “environment”.

• Partially canceling / invalidating means deliberately undermining or rejecting fixed definitions of “computation” and “environment”, refusing to treat them as rigid, stable entities.

Open Computing overlaps these two operations:

• It connects computation and environment so they influence and reshape each other.

• It also refuses to let their old, clear-cut definitions stand.

The overlap produces a dynamic state where boundaries are constantly renegotiated.
However, it is very difficult to implement open computing in programmable computation, since it refers to the mixture within a logical framework and the external denial of the framework's assumptions. These are different logical hierarchies, or levels of cognition. It can be analogous to the coexistence of self-reference and the frame problem as this metaphor \cite{gunji2008abstract}. A self-referential sentence such as "This sentence is false" results in a vicious cycle or contradiction, assuming that "this," "sentence," and "false" are self-evident. However, when the frame problem is introduced, it leads to a situation where the meaning of "this," "sentence," and "false" becomes unclear, the contradiction itself does not even exist, and the system no longer cares logic and/or programmable computation, leading to the unconventional computation deviating from the framework of programmable computation.

With these caveats in mind, we can say the following about the calculation process at the material level: Because there is no rigid separation, the system can integrate perturbations (unexpected inputs, environmental changes, noise) more fluidly. This allows for adaptability and robustness. A traditional computing system is like a train running on fixed tracks (software = train, environment = tracks). Open Computing is more like a boat in a river — the vessel and the water influence each other’s motion, and the boundary between “navigation” and “environment” is fluid.

Scientific Parallels Beyond the Boat Analogy are list up by the following.
In quantum computing \cite{feynman2018simulating, deutsch1985quantum, arute2019quantum, preskill2023quantum}, the algorithm and the physical qubits are not strictly separate, since the environment (noise, de-coherence) directly shapes the computation and the computation (error correction, adaptive measurement) actively reshapes how the environment is handled. Results emerge from their entangled interplay. Analog computers \cite{small2013analogue, Adamatzky2022} solve equations through the continuous evolution of voltages and currents. The physical properties of the hardware are the computation. Environmental factors like temperature directly affect outcomes. Neuromorphic systems \cite{mcculloch1943logical, hopfield1982neural, indiveri2015memory, markovic2020physics} mimic biological neural networks, with computation happening in physical substrates that adapt over time. Hardware and computation co-evolve, blurring the separation between program and environment.

\section{Open computing in chemical reaction}\label{sec3}

\subsection{Type and Token computing in chemical reaction}\label{subsec1}

Here, the relationship between the computation and the computation execution environment is implemented as open computing, but we specifically implement it in a dual form of Token computing and Type computing. This is because, as we are considering chemical reactions as computations, the issue of the buffer between the computation and the computation execution environment becomes an issue for both microscopic systems of particles and macroscopic systems that define concentration. When making a distinction between microscopic and macroscopic, it often becomes an issue of molecules and concentration, but here we consider it with respect to the context of semiotics, and following the concepts of type and Token used in semiotics, we envision a computational implementation called Type computing and Token computing, and by the interplay of both, we implement open computing for chemical reactions.

    In semiotics, we believe that the way we perceive everything is determined by the way we perceive it through experience. The plastic bottle that we imagine in our minds to hold liquid is of course a symbol, but the plastic bottle that I am holding to drink water is also a symbol. It is a certain substance, but because it is recognized as a plastic bottle, this plastic bottle that I am holding is also a symbol. Even if the symbols are the same, the symbol that corresponds to the general idea of a plastic bottle is called a type, and the symbol that is recognized as an individual is called a Token. Therefore, although it is a relative term, it is fair to say that types are virtual rather than material, and Tokens are material rather than virtual. A type can be thought of as a norm that encompasses the diversity that Tokens show.
    
    Particularly in chemical reactions, Token computing is defined based on molecules for both the computation and the computation execution environment. Computation is the activation and inactivation of molecules in a chemical reaction, and the computation execution environment is defined by the size of the molecular aggregate that determines the rate of activation and inactivation. In contrast, the Type computing of a chemical reaction is given as a norm that encompasses the entire activation and execution environment. In fact, Token computing and Type computing in chemical reactions are defined as follows.

\subsection{Token computing in chemical reaction}\label{subsec2}

Token computing is a computational model of chemical reactions based on molecular behavior. The issues we are concerned with here are chemical reactions as computational processes and the microscopic reaction environment as the computational environment. As mentioned above, when defining the concentration of chemical species and constructing a model with differential equations, it is assumed that molecules are evenly distributed in the solution. However, in chemical reactions within cells, the molecular density is highly biased, and various aggregation states are possible, such as monomers, the formation of clusters of various sizes, and the formation of giant clusters. Furthermore, the rate of chemical reactions changes significantly depending on the degree of molecular aggregation.

\begin{figure}[h]
    \centering
    \includegraphics[width=1.0\linewidth]{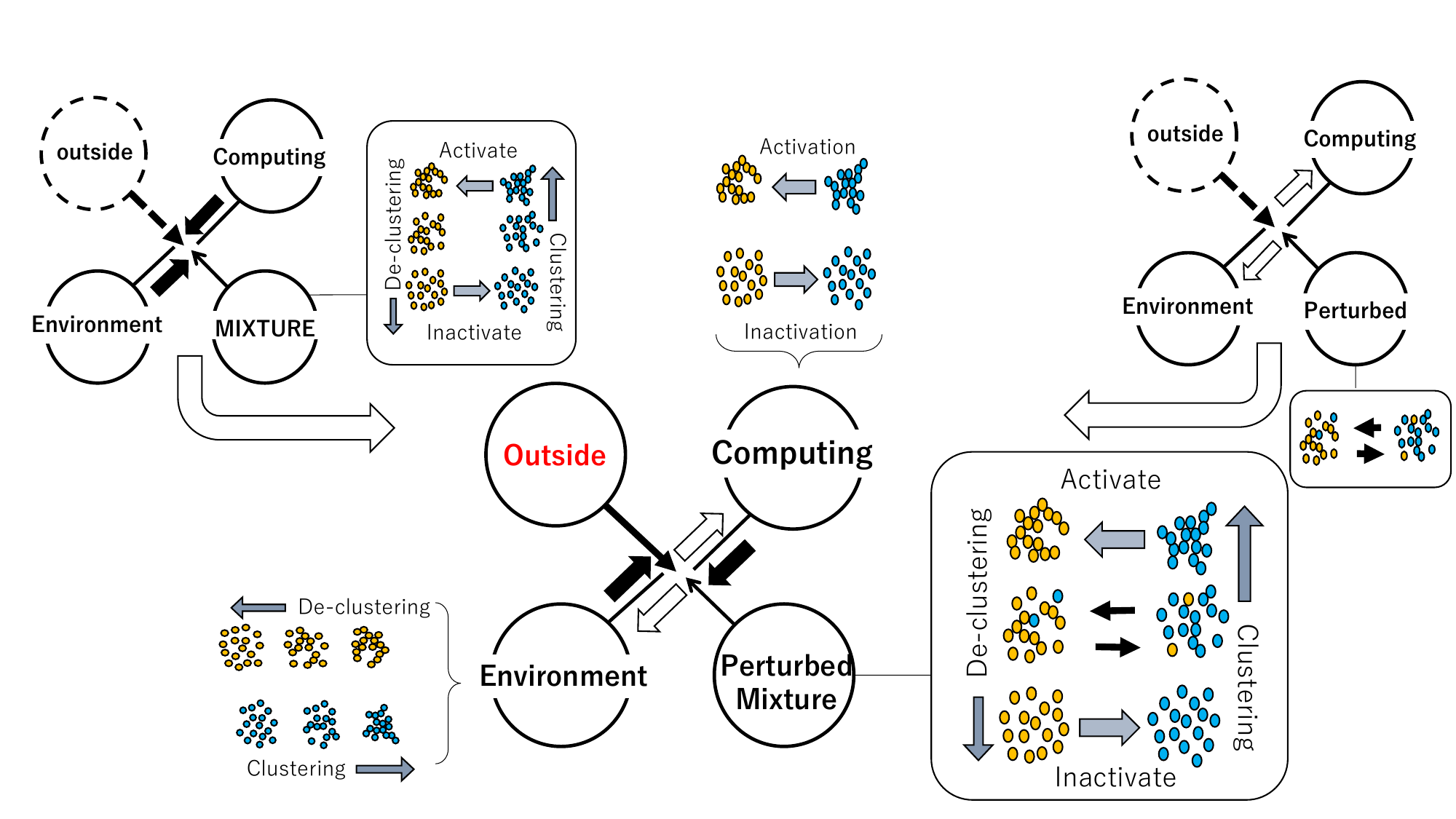}
    \caption{Schematic diagram of Open computing in chemical reaction, featuring co-existence of mixture and invalidation of computing and its execution environment. Computing is defined by activation and inactivation of molecules, and its execution environment is defined by clustering and de-clustering. Invalidation of them is implemented by introducing perturbation. See text for the detail.}
    \label{fig:enter-label}
\end{figure}

Taking the above into consideration, here we define the toy model of chemical reactions as follows. First, a chemical reaction as a computation is defined only in terms of the activation and inactivation of molecules. Second, the computational environment is given in terms of the molecular aggregation state. In other words, the computational environment is defined in terms of the clustering and de-clustering of molecules, which changes the molecular aggregation state. It is assumed that activation and inactivation change depending on the degree of molecular aggregation, which makes it possible to envision the interaction between the computation and the computational environment.

Of course, the above computation is implemented as open computing. As shown in the lower center of Fig 2, computing (chemical reaction) is defined as inactive molecules changing into active molecules, which are monomers, and as active molecules changing into inactive molecules when the cluster is at its maximum (a state in which all molecules are aggregated). Here, blue circles represent inactive molecules and yellow circles represent active molecules. In contrast, the computing execution environment (clustering and de-clustering of molecules) is defined as follows: de-clustering proceeds when the ratio of active molecules in the cluster constituent molecules is greater than 0.5, and clustering proceeds when it is not. Therefore, in an ideal state, active molecules proceed with de-clustering and inactive molecules proceed with clustering. The above is expressed as molecules as Tokens in the following equations.

Assume that there are \(N\) molecules of a certain chemical species. Some molecules aggregate to form a cluster. If all molecules are monomers, the number of clusters is \(N\), and if all molecules are aggregated, the number of clusters is 1. Here, the molecules are numbered from 1 to \(N\). Since molecules aggregate to form clusters, the clusters at time \(t\) are also numbered, but when counting as clusters, the monomers are also called clusters consisting of one molecule, and the clusters are numbered from 1 to \(C_\text{max}^t\). Here, \(C_\text{max}^t\) represents the total number of clusters at time \(t\). The two characteristics of molecule \(k\) at time \(t\) are represented by two variables, \(m_0^t(k)\) represents the number of the cluster to which molecule \(k\) belongs, \(m_0^{t}(k)\) \(\in\)\{1, 2, …, \(C_\text{max}^t\)\}. The variable \(m_1^{t}(k)\) indicates whether the molecule is active (1) or inactive (0), \(m_1^{t}(k)\) \(\in\) \{0, 1\}.

Similarly, cluster \(n\) at time \(t\) is represented by two variables, \(C_0^t(n)\) represents the number of molecules that make up this cluster, \(C_0^t(n)\) \(\in\) \{1, 2, …, \(N\)\}. \(C_1^t(n)\) represents the number of active molecules in this cluster, \(C_1^t(n)\) \(\in\) \{0, 1, …, \(C_0^t(n)\)\}. Molecules \(s\) in cluster \(n\) are also numbered, \(s\) \(\in\) \{0, 1, …, \(C_0^t(n)\)\}, and \(Cl^t(n, s)\) indicates the serial number of the \(s\)th molecule in the cluster, \(Cl^t(n, s)\) \(\in\) \{1, 2, …, \(N\)\}.

Using these variables, we define clustering and de-clustering as follows. First, clustering is calculated as follows. First, two clusters \(p\) and \(q\) are randomly selected. Let \(\text{Rand}(n)\) be a function that selects a natural number uniformly at random from 1 to \(n\), and \(p\) and \(q\) are determined so that \(p<q\) with the selected value.

\begin{equation}
    p=\text{Rand}(C_\text{max}^t), 
    q=\text{Rand}(C_\text{max}^t), 
    p<q.
\end{equation}

However, cluster \(p\) and \(q\) must satisfy the following clustering conditions, where \(0.0\leq\theta_c\leq1.0\). 

\begin{equation}
C_1^t(p)/ C_0^t(p)<\theta_c, C_1^t(q)/ C_0^t(q)<\theta_c
\end{equation}

 Since these two clusters are fused to form a new composite cluster, the number of the new composite cluster is set to the smaller number, that is, \(p\) since \(p<q\), and the number of molecules and the number of active molecules are assigned to this new composite cluster.

\begin{equation}
 C_0^{t+1}(p)=C_0^t(p)+C_0^t(q)
\end{equation}
\begin{equation}
 C_1^{t+1}(p)=C_1^t(p)+C_1^t(q)
\end{equation}

The molecules that make up the new synthetic cluster are as follows.
\begin{equation}
Cl^{(t+1)}(p, x)=Cl^t(p, x), x=1, 2,..., C_0^t(p),
\end{equation}
\begin{equation}
Cl^{t+1}(p, y+C_0^t(p))=Cl^t(q, y), y=1, 2, ..., C_0^t(q).
\end{equation}

In line with the above clustering, new cluster numbers are assigned to the molecules that belong to the new synthetic cluster. That is,

\begin{equation}
 m_0^{t+1} (z)=p, z= Cl^{t+1}(p, x), x=1, 2, ..., C_0^t(p)+ C_0^t(q)   
\end{equation}

The address of the \(q\)th cluster will be freed by the clustering synthesis, so the numbers will be packed.

\begin{equation}
C_0^{t+1}(s)=C_0^t(s+1), s=q, q+1,..., C_\text{max}^t
\end{equation}
\begin{equation}
C_1^{t+1}(s)=C_1^t(s+1), s=q, q+1,..., C_\text{max}^t
\end{equation}

Similarly,

\begin{equation}
Cl^{t+1}(s, x)=Cl^t(s+1, x), s=q, q+1,..., C_\text{max}^t, x=1, 2,..., C_0^t(s)
\end{equation}

Furthermore, the cluster numbers of the monomers belonging to the cluster are also filled in, if \(m_0^t(s)\ge q\) with \(s=1, 2, \dots, N\).

\begin{equation}
m_0^{t+1}(s)=m_0^t(s)-1, 
\end{equation}

For all clusters not calculated by the above procedure, i.e., for \(k \in \{1, 2, \dots, C_{\text{max}}^t\} \setminus \{p, q, q+1, \dots, C_{\text{max}}^t\}\),

\begin{equation}
C_0^{t+1}(k)=C_0^t(k)   
\end{equation}

\begin{equation}
C_1^{t+1}(k)=C_1^t(k)  
\end{equation}

For a molecule \(s\) that belongs to such a cluster \(k\), i.e., for a molecule \(s\) such that \(m_0^t(s)=k\)

\begin{equation}
   m_0^{t+1}(s)= m_0^t(s) ,
\end{equation}
\begin{equation}
   m_1^{t+1}(s)=m_1^t(s) .
\end{equation}
\begin{equation}
 Cl^{t+1}(k, r)=Cl^t(k, r), r = 1, 2, …, C_0^t(k).   
\end{equation}

Next, we define the de-clustering process. First, we select a cluster that will be de-polymerized and split into two. We select a monomer uniformly at random so that it is selected depending on the size of the cluster, and select a cluster k that contains that monomer. In other words, there exists \(p\) in \{1, 2,..., \(C_0^t(k)\) \} such that

\begin{equation}
  Cl^t(k, p) = \text{Rand}(N).  
\end{equation}

Here, we check that the selected cluster satisfies the following conditions, where \(0.0\leq\theta_{dec}\leq1.0\).
\begin{equation}
 C_1^t(k)/ C_0^t(k)>\theta_{dec}  
\end{equation}

Randomly select molecule \(s\) from the molecules that make up cluster \(k\)
\begin{equation}
   s=\text{Rand}(C_0^t(k)-1) 
\end{equation}

Here, \(C_0^t(k)-1\) is used instead of \(C_0^t(k)\) because the cluster is divided into two before and after \(s\), and if \(s=C_0^t(k)\) it is not divided. The two newly obtained clusters are cluster \(k\) and the other is the last cluster, and are called cluster \(C_{\text{max}}^t+1\). At this time, the number of molecules that make up each cluster is

\begin{equation}
C_0^{t+1}(k)=s,
\end{equation}

\begin{equation}
 C_0^{t+1}(C_\text{max}^t+1)=C_0^t(k)-s.   
\end{equation}

The number of active molecules in each divided cluster is counted and determined as follows:

\begin{equation}
  C_1^{t+1}(k)=\#\{j \in \{1, 2,\dots, s\} \mid m_1^t(Cl^t(k, j))=1\}   
\end{equation}
\begin{equation}
   C_1^{t+1}( C_\text{max}^t+1)=\#\{j \in \{1, 2, \dots, C_0^t(k)-s \} 
 \mid m_1^t(Cl^t(C_\text{max}^t+1, j))=1\} 
\end{equation}

Also, in cluster \(k\), for \(j=1, 2, \dots, s\),
\begin{equation}
Cl^{t+1}(k, j)= Cl^t(k, j),
\end{equation}
\begin{equation}
m_0^{t+1}( Cl^t(k, j))= m_0^t(Cl^t(k, j))
\end{equation}
\begin{equation}
m_1^{t+1}( Cl^t(k, j))= m_1^t(Cl^t(k, j))
\end{equation}

Similarly, for cluster \(C_\text{max}^t+1\), for \(j=1, 2, \dots, C_0^t(k)-s\),
\begin{equation}
Cl^{t+1}(C_\text{max}^t+1, j)= Cl^t(C_\text{max}^t+1, j+s),
\end{equation}
\begin{equation}
m_0^{t+1}(Cl^t(C_\text{max}^t+1, j))= m_0^t(Cl^t(C_\text{max}^t+1, j+s)) 
\end{equation}
\begin{equation}
 m_1^{t+1}(Cl^t(C_\text{max}^t+1, j))= m_1^t(Cl^t(C_\text{max}^t+1, j+s)).   
\end{equation}

For all clusters not calculated by the above procedure, that is,
\( m \in \{1, 2, \dots, C_{\text{max}}^t\} \setminus \{k\} \)

\begin{equation}
 C_0^{t+1}(m)=C_0^t(m)   
\end{equation}

\begin{equation}
  C_1^{t+1}(m)=C_1^t(m)  
\end{equation}

For a molecule \(s\) that belongs to such a cluster $m$, that is, a molecule $s$ such that \(m_0^t(s)=m\)
\begin{equation}
m_0^{t+1}(s)=m_0^t(s)
\end{equation}
\begin{equation}
m_1^{t+1}(s)=m_1^t(s)  
\end{equation}
\begin{equation}
 Cl^{t+1}(m, r)=Cl^t(m, r), r=1, 2,\dots, C_0^t(m).   
\end{equation}

Finally, we define activation and inactivation. When all molecules become monomers, inactivation occurs in all molecules. In other words, if \(C_\text{max}^t=N\), then for any molecule \(i\)
\begin{equation}
 m_1^{t+1}(i)=0, 
\end{equation}
\begin{equation}
 C_1^{t+1}(i)=0.   
\end{equation}

Also, when all \(N\) molecules aggregate and become one giant cluster, all molecules become activated. In other words, if \(C_\text{max}^t=1\) then for any molecule \(i\)

\begin{equation}
m_1^{t+1}(i)=1, 
\end{equation}
\begin{equation}
C_1^{t+1}(1)=N.  
\end{equation}

Now that the computing and execution environment for chemical reactions have been defined, the first step is to implement the confusion between the two. This is achieved by connecting the computing and computing execution environment in an ideal state which is without perturbation (top left of Fig 2). In an ideal state, activation proceeds all at once in the maximum cluster, and active molecules proceed with de-clustering and become monomers. Active monomers are transformed into inactive monomers at once and inactive molecules proceed to cluster. This process proceeds in a circular manner. This cycle is very confusing, caused by the connection between computing and the execution environment for computing.

Secondly, the invalidation of computing and the computing execution environment in chemical reactions is defined. Here, computing and the computing execution environment are envisioned in an ideal state, so clustering is defined by inactive molecules, and de-clustering is defined by active molecules (top right of the lower center diagram in Fig 2). Computing is also defined by activation in the largest cluster and inactivation in monomers. Therefore, if even a small amount of noise is introduced here, these definitions are disturbed, and the definition in the strict sense is invalidated. With a certain small probability, active molecules are inactivated, and inactive molecules are activated. This perturbation is defined by the transition with probability \(0.0\leq P_\text{noise}\leq1.0\)

   \begin{equation}
       m_0^{t+1}=(m_0^t+1)\%2
   \end{equation}
      
This noise becomes the invalidation of computing and the computing execution environment (top right of Fig 2).
The superposition of computing and computing execution environment in chemical reactions becomes open computing in Token computing of chemical reactions (bottom center of Fig 2). This system makes it possible to define invalidation only by introducing small fluctuations, so it is essential to introduce fluctuations from outside. In other words, as shown in Fig 2, the mixture of computing and its execution environment and the superposition of invalidation constantly recruit fluctuations outside the system.

\subsection{Type computing in chemical reaction}\label{subsec3}

\subsubsection{Type as binary relation}\label{subsubsec1}

Type computing in chemical reactions does not distinguish between assumed chemical reactions with respect to the time series of molecular behavior, but expresses them in terms of the relationship between clusters and de-clusters in the normative sense and between active and inactive molecules. Basically, it takes the form of a transition matrix of molecular states. As shown in Fig 3, the transition diagram is expressed as a matrix with mixture and invalidation in open computing as sub-matrices. Since this matrix is a transition matrix, the \((i, j)\) element indicates the probability that the state in row \(i\) transitions to the state in column \(j\). A blank cell indicates that the transition probability is 0. The darker the cell, the higher the probability, and it is assumed that the difference in probability between the dark blue grid and the light blue grid is sufficiently large, and in the mixture phase, in most cases, the transition is realized only in the 'diagonal sub-matrix' (a part of a diagonal matrix that contains diagonal elements).

\begin{figure} [h]
    \centering
    \includegraphics[width=1\linewidth]{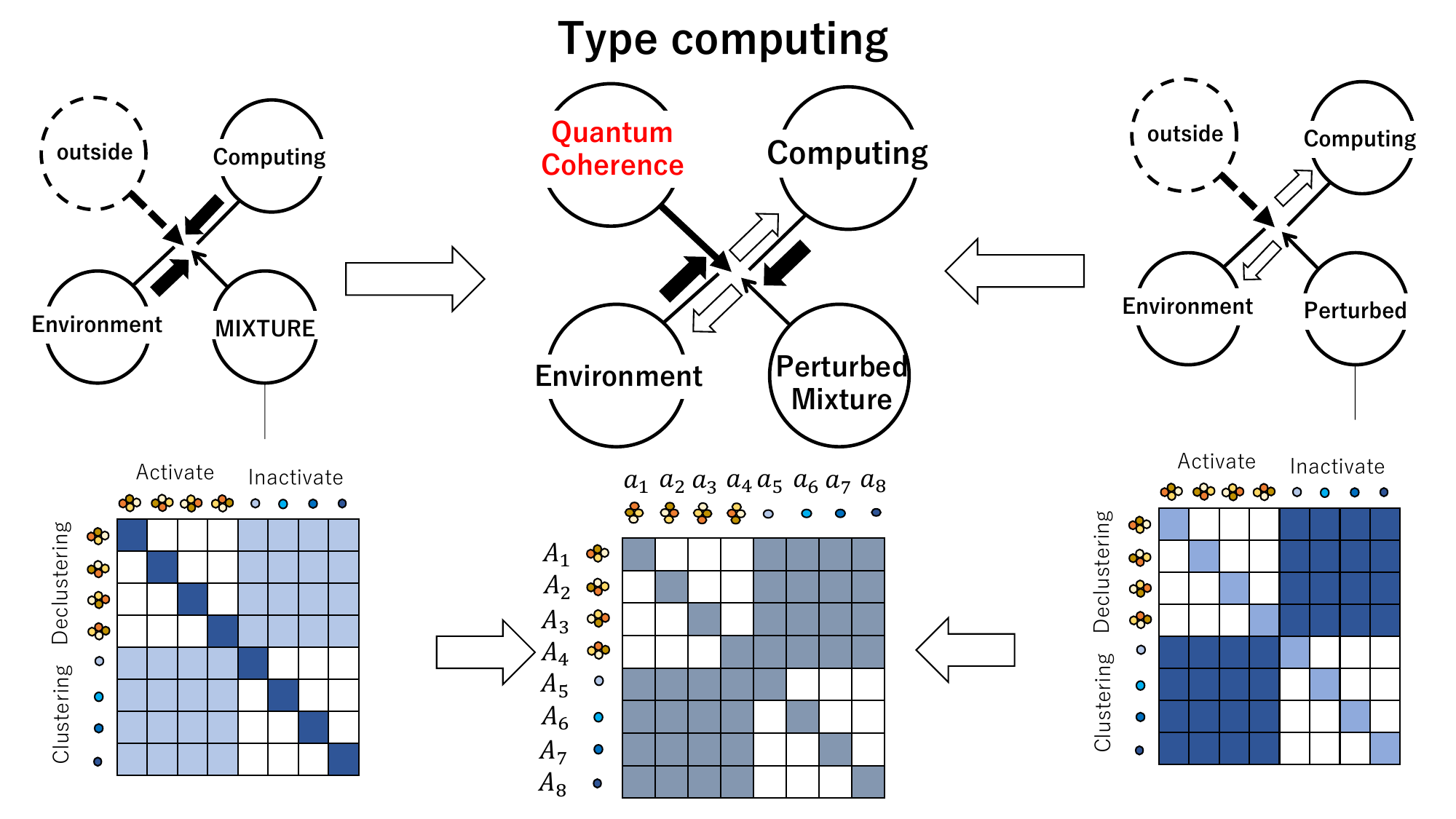}
    \caption{Schematic diagram of Type computing in chemical reaction. The transition matrices below show transition in from \(t\)th step to \(t+1\)th step. Darker color cells represent the transition with higher probability. Warm and cool color circles represent active and inactive molecules, respectively. A left matrix shows the clustering in inactive molecules and de-clustering in active molecules (mixture of computing and its environment). A right matrix shows the random transition from active to inactive molecules and vice versa. Superposition of the left and right matrices yields a binary relation located in center.}
    \label{fig:enter-label}
\end{figure}

The toy model of the chemical reaction that is realized is the same as that assumed in Token computing. The lower left diagram in Fig 3 shows the mixture of computing and its execution environment in Type computing. The diagonal sub-matrix represents the clustering of inactive molecules and the de-clustering of active molecules assumed in Token computing. The upper half of the vertical axis represents active molecules, the lower half represents inactive molecules, and the left half represents active molecules and the right half represents inactive molecules. The circles arranged on the sides of the matrix represent molecules, with the warm-colored circles representing active molecules and the cool-colored circles representing inactive molecules. The upper left diagonal sub-matrix shows de-clustering specialized for active molecules, and the lower right diagonal sub-matrix shows clustering specialized for inactive molecules. Here, only four molecules are considered, and the aggregate in which all of them are bonded is the largest cluster. The four molecules are distinguished from each other and are colored by shades of color, not limited to warm and cool colors. As will be described later, the numerical simulations are performed with 200 molecules. Furthermore, the molecular aggregation state is merely a convenient way of indicating clustering and de-clustering. For example, it does not only indicate de-clustering from 4 to 3, but it is assumed that de-clustering and clustering of all numbers are realized by this transition matrix.

The invalidation of computing and its environment is represented by the upper right and lower left "relational sub-matrices” (sub-matrices with non-zero probability in all cells). Considering that the size of the molecular aggregation is a convenient one, it means the exchange of active and inactive molecules independent of the size of aggregates. However, since the upper half of the matrix represents active molecules and the lower half represents inactive molecules, there is a difference between the transition in which an active molecule changes to an inactive molecule and the transition in which an inactive molecule transitions to an active molecule.

The transition probability in simulating studies is obtained by calculating the frequency distribution in all clusters, selecting the cluster with the largest number, and determining the overall behavior depending on whether the active molecule rate in that cluster is 0.5 or higher. In other words, if it is 0.5 or higher, active molecules are inactivated, and in other cases, inactive molecules are activated. The mechanism by which this is realized for all molecules will be described later. Here, it is sufficient to confirm that the mixture of computing (activation/deactivation) is shown in the diagonal sub-matrices, and the invalidation of its execution environment (clustering/de-clustering) is shown in the relational sub-matrix.
   
Open Computing as Type computing is realized by superimposing mixture and invalidation. The lower center Fig of Fig. 3 shows two matrices showing superimposed transition probabilities. By averaging high-probability transitions and low-probability ones, the transition probability becomes either present or absent (0), which is a binary relation. That is why each cell is shown as either gray or blank. The transition matrix (i.e., binary relation) overlapping mixture and invalidation shows the situation shown in Token computing, from activation to de-clustering by activated molecules, inactivation, and subsequent clustering by inactive molecules, with fluctuations in the vibration.

\subsubsection{Quantum logic in rough set approximation}\label{subsubsec2}

Using the rough set technique from the binary relation, an algebraic structure called a lattice can be obtained. Here,  vertical and horizontal are the same set \(S\), and \(I\subseteq S\times S\) is a binary relation. However, what is lined up vertically are equivalence classes of equivalence relation \(R\) in \(S\), and what is lined up horizontally are equivalence classes of equivalence relation \(K\) in \(S\).

An equivalence class of \(x \in S \) with respect to an equivalence relation \(R\) on \(S\) is denoted by \(A_1, A_2, \dots, A_8\), and is defined as:
\begin{equation}
    A_k = \{y\in S \mid xRy \},  k=1, 2, \dots, 8.
\end{equation}
Similarly, an equivalence class of \(x \in S \) with respect to another equivalence relation \(K\) on \(S\) is denoted by \(a_1, a_2, \dots, a_8 \), and is defined as:
\begin{equation}
    a_k = \{y\in S \mid xKy \}, k=1, 2, \dots, 8.
\end{equation}

Note that an equivalence class with respect to an equivalence relation \(R\) is typically denoted by \([x]_R\), and that the union of all equivalence classes forms the set \(S\):
\begin{equation}
    \bigcup_{k=1}^{8}A_k=\bigcup_{k=1}^{8}a_k=S.
\end{equation}

Now, define a binary relation \(I\) by:
\begin{equation}
    A_iIa_j   \iff \exists x \in S (x \in A_i \land x \in a_j)
\end{equation}

A lattice --- an ordered set closed with respect to join and meet --- can be obtained from a binary relation \(I\) using the rough set approximation framework. 

Given an equivalence relation \(J\), for any subset \(X \subseteq S\), the lower approximation of \(X\) is defined as:
\begin{equation}
    J_* (X)=\{y\in S \mid [x]_J \subseteq X \},
\end{equation}

and the upper approximation of \(X\) is defined as:
\begin{equation}
    J^* (X)=\{y\in S \mid [x]_J \cap X \ne \varnothing \},
\end{equation}

Using these approximations, we define a lattice as:
\begin{equation}
    L=\{X \subseteq S \mid R_*(K^*(X))=X\}
\end{equation} 

The composition \(R_*(K^*(-))\) is referred to as a closure operator.
The lower and upper approximations can be re-expressed using the binary relation \(I\). When we replace the set \(S\) with \(S_1=\{A_1, A_2, \dots, A_8\}\) or \(S_2=\{a_1, a_2, \dots, a_8\}\), the upper approximation for \(X \subseteq S_1\) can be rewritten as:
\begin{equation}
    K^*(X)=\{q \in S_2 \mid pIq, p \in X \}.
\end{equation}
Similarly, the lower approximation for \(Y \subseteq S_2\) is given by
\begin{equation}
    R_*(Y)=S_1 \setminus \{p \in S_1 \mid pIq, q \notin Y \}.
\end{equation}

Ultimately, we obtain a non-distributive orthomodular lattice --- commonly known as Quantum logic --- from the binary relation \(I\) shown in Fig. 3. We refer to this structure as a quantum lattice, which is composed of multiple Boolean sub-lattices that share certain elements. 

To better understand the quantum lattice, we first revisit the concept of a lattice and a Boolean lattice (Boolean algebra) \cite{davey2002introduction, gunji2022psychological, gunji2010non, gunji2016quantum}. 

As previously mentioned, a lattice is an ordered set closed under binary operations called join and meet. The elements of the lattice derived from the condition \(R_*(K^*(X))=X\) are subsets of \(S_1\)and the ordering is defined by set inclusion.

\begin{figure}
    \centering
    \includegraphics[width=1\linewidth]{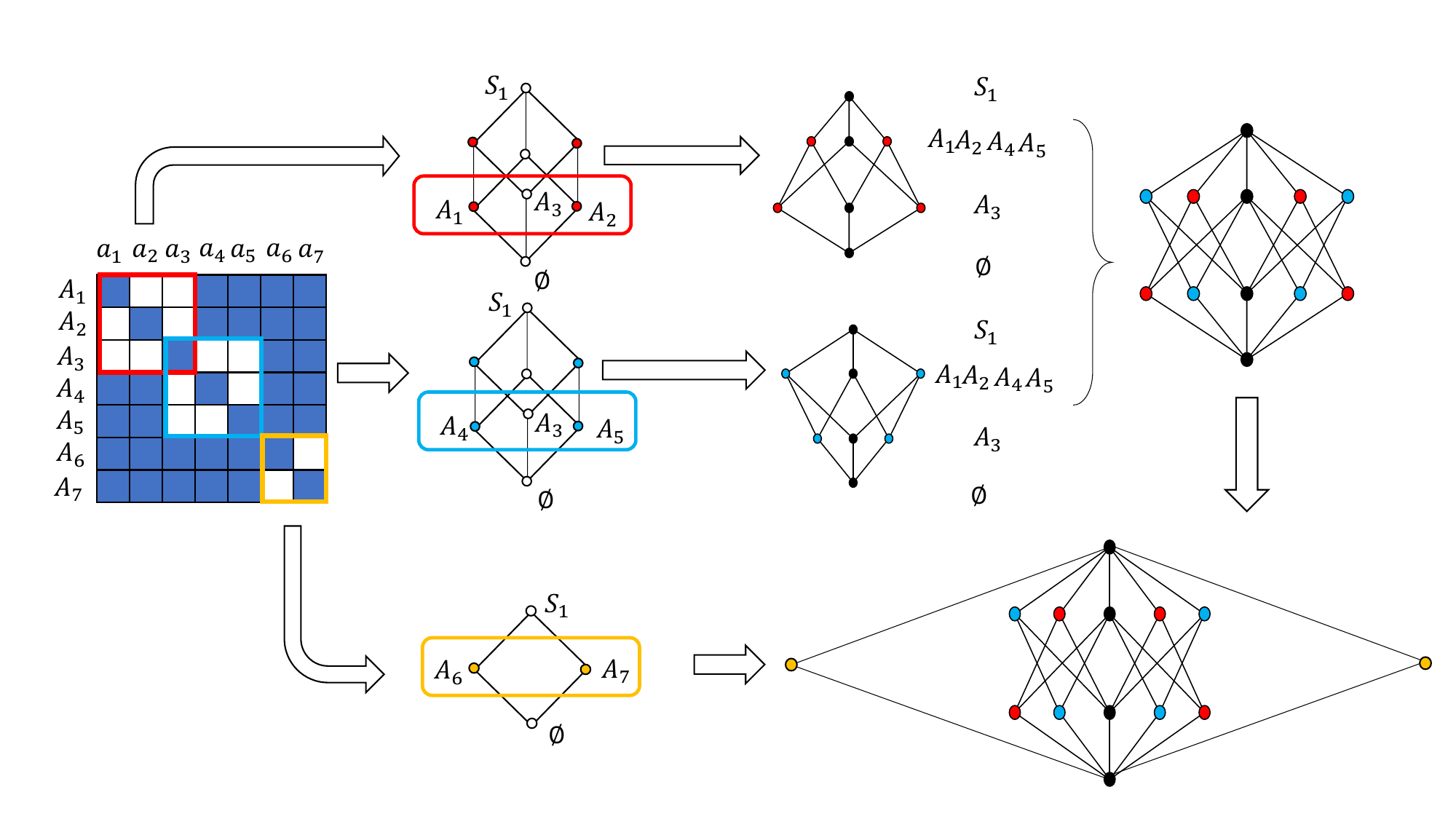}
    \caption{A lattice is constructed from a binary relation using the closure operator derived from rough set approximations. The binary relation (left) is partitioned into two overlapping \(3 \times 3\)diagonal sub-relations and one \(2 \times 2\)diagonal sub-relation, which is surrounded by related cells (indicated in blue). Blank cells represent the absence of a relation. Applying the closure operator defined by rough set approximations yields a lattice structure. Each diagonal sub-relation individually generates a Boolean lattice, represented by the Hasse diagram (center). The overlapping sub-relations and the related surrounding cells give rise to lattice elements that are shared among multiple Boolean sub-algebras. A lattice formed by several Boolean sub-algebras sharing common elements constitutes a non-distributive orthomodular lattice—i.e., a structure corresponding to quantum logic.}
    \label{fig:enter-label}
\end{figure}

Fig. 4 illustrates an example of quantum lattice. The binary relation \(I\) consists of diagonal sub-relations, depicted by red, blue and orange loops. The read and blue sub-relations overlap, and each diagonal sub-relations are surrounded by related cells. A relation \(R\) of size \(n\times n\) in which only diagonal elements are related and i.e., \(iRi\) is called a diagonal relation. Assume that the red-loop sub-relation in \(I\) in Fig. 4 is isolated, forming a diagonal relation on \( \subseteq \{A_1, A_2, A_3\} \times \{a_1, a_2, a_3\}\). In this case, it can be easily verified that:

\begin{equation}
R_*(K^*(\{A_1\}))=R_*(a_1)=A_1.
\end{equation}

Since for any \(X \subseteq \{A_1, A_2, A_3\}\), we have \(R_*(K^*(\{X\}))=X\), the resulting lattice is:

\begin{equation}
  L=\{\varnothing, \{A_1\}, \{A_2\},\{A_3\}, \{A_1, A_2\},\{A_2, A_3\},\{A_1, A_3\},\{A_1, A_2, A_3\}\}  
\end{equation}
 forms a Boolean lattice. A power set is a lattice ordered by subset inclusion. The meeting of two elements of a lattice, \(x, y \in L\) denoted  \(x \land y\) is defined by:
\begin{equation}
    x \land y\leq x, x \land y\leq y
\end{equation}
\begin{equation}
    z\leq x, z\leq y \Rightarrow z \leq x \land y.
\end{equation}

The Join \(x \lor y\) is similarly defined by replacing \(\leq\) with \(\geq\) and \(x \land y\) with \(x \lor y\).

A lattice can be visualized using a Hasse diagram. If two elements \(x, y (x<y)\) have no element between them, they are connected by a line drawn with \(x\) below \(y\). The power set of 
 \(\{A_1, A_2, A_3\}\) is shown in top-left Hasse diagram in Fig. 4. The atoms --- elements immediately above the empty set (the least element) --- are labeled by \(A_k\) instead of \(\{A_k\}\) for simplicity. For example, it is clear that \(\{A_1, A_2\}\land \{A_2, A_3\}=\{A_2\}\). Hence for any elements \(x, y\) in any Boolean lattice
\begin{equation}
    x \land y = x \cap y,
\end{equation}
 \begin{equation}
    x \lor y = x \cup y.
\end{equation}

In this sense, Boolean lattice corresponds to classical logic.
For a \(3 \times 3\) diagonal relation, we obtain a Boolean algebra with \(2^3\)elements. This can be generalized, and for an \(n\times n\) diagonal relation, we obtain a Boolean algebra with \(2^n\) elements. So, what happens in the case of a diagonal relation surrounded by relation elements as in Fig. 4? Now, if we calculate the closure operation using the entire \(8\times8\) relation shown in Fig. 4, we obtain

\begin{equation}
 R_*(K^*(\{A_1\}))=R_*(\{a_1, a_4, a_5, a_6, a_7\})=\{A_1\}   
\end{equation}

which is an element of the lattice. All subsets of \(\{A_1, A_2, A_3\}\) are fixed points for the closure operation, which means that the lattice formed by the entire \(8\times8\) relation contains a Boolean algebra with \(2^3\) elements. The same is true for the subsets \(\{A_3, A_4, A_5\}\). We can also see that the subset \(\{A_6, A_7\}\)is a Boolean algebra with \(2^2\) elements.

What happens to the duplicated elements of the diagonal relation? In the diagonal relation \(\{A_1, A_2, A_3\}\times\{a_1, a_2, a_3\}\) and the diagonal relation \(\{A_3, A_4, A_5\}\times\{a_3, a_4, a_5\}\) overlap. If we calculate the closure operation using the entire \(8\times8\) relation, we get

\begin{equation}
  R_*(K^*(\{A_3\}))= R_*(\{a_3, a_6, a_7\})=\{A_3\}  
\end{equation}

This shows that the two Boolean algebras have something in common. So, if we calculate all the closure operations using the entire \(8\times8\) relation for \(\{A_1, A_2, A_3, A_4, A_5\}\) related to the two diagonal relations, we get

\begin{equation}
  R_*K^*(\{A_1, A_2\})= R_*(\{a_1, a_2, a_4, a_5, a_6, a_7\})=\{A_1, A_2,A_4, A_5\}  
\end{equation}
\begin{equation}
  R_*K^*(\{A_4, A_5\})= R_*(\{a_1, a_2, a_4, a_5, a_6, a_7\})=\{A_1, A_2,A_4, A_5\}  
\end{equation}
\begin{equation}
  R_*K^*(\{A_1, A_2, A_4, A_5\})= R_*(\{a_1, a_2, a_4, a_5, a_6, a_7\})=\{A_1, A_2,A_4, A_5\}  
\end{equation}

And we can see that \(\{A_1, A_2, A_4, A_5\}\) is the second common element of the two Boolean algebras. Moreover,

\begin{equation}
R_*K^*(S_1)= R_*(S_2)=S_1
\end{equation}
\begin{equation}
R_*K^*(\varnothing)= R_*(\varnothing)= \varnothing.
\end{equation}

The number of overlapping elements of two Boolean algebras with \(2^3\) elements is four, and it is shown as the Hasse diagram in the upper center of Fig. 4. The structure where two Boolean algebras with \(2^3\) elements overlap is no longer a Boolean algebra, but a non-distributive orthomodular lattice (i.e., quantum lattice).

Here, if we calculate the closure operation for another diagonal relation, \(\{A_6, A_7\}\times\{a_6, a_7\}\), on the entire \(8\times8\) relation,

\begin{equation}
    R_*K^*(S_1)= R_*(S_2)=S_1,
\end{equation}
\begin{equation}
    R_*K^*(\varnothing)= R_*(\varnothing)=\varnothing,
\end{equation}
\begin{equation}
    R_*K^*(\{A_6\})= R_*(\{a_6\})=\{A_6\},
\end{equation}
\begin{equation}
    R_*K^*(\{A_7\})= R_*(\{a_7\})=\{A_7\}.
\end{equation}

It can be seen that it is a Boolean algebra with \(2^2\) elements (Hasse diagram in the lower left of Fig. 4). However, the greatest and least elements overlap with the quantum lattice mentioned above, so if we draw a Hasse diagram by overlapping them, we obtain a lattice shown in the Hasse diagram in the lower right of Fig. 4. This is also a quantum lattice.

From the above, the characteristics of quantum lattice can be described as follows. First, they have multiple Boolean algebras as parts. These are called Boolean sub-lattices. Second, multiple Boolean sub-algebras share at least the least and greatest elements, and even if they share more elements, they do so in a way that satisfies the orthomodular law.

In this case, a quantum lattice can be compared to quantum mechanics as follows. Each Boolean sub-lattice can be thought of as a different Hilbert space. The atoms of a partial Boolean lattice (the elements directly above the least element) correspond to the basis vectors of the Hilbert space. Superposition is represented by a union, since the join is a union. The existence of multiple partial Boolean lattices represents a composition of different Hilbert spaces. Therefore, the elements of the lattice shared by different Boolean sub-lattices correspond to quantum entanglement. In this way, quantum lattices allow us to consider quantum entanglement, interactions between different Hilbert spaces, and even quantum coherence.

The important point is that if we take elements from different diagonal sub-relations \(\subseteq S_1\), it will not be an element of the lattice. In fact, if we apply the closure operator to \(\{A_1, A_6\}\), which is a set of elements selected from the diagonal relations of \(\{A_1, A_2, A_3\}\) and the diagonal relations of \(\{A_6, A_7\}\), we get

\begin{equation}
    R_*(K^*(\{A_1, A_6\}=R_*(S_2)=S_1.
\end{equation}

It is not a fixed point for the closure operator, so it is not an element of the lattice. Therefore, in the different Boolean sub-algebras obtained from the \(7\times 7\) binary relations shown in Fig. 4, there are no shared elements other than those mentioned previously, and the sub-algebras are separable except by entanglement.

\subsubsection{Quantum logic in chemical reaction}\label{subsubsec3}

Let us now return to the lattice obtained from the binary relation obtained in Type computing in chemical reaction in Fig. 3. The \(8 \times 8\) relation has two \(4 \times 4\) sub-diagonal relations, which are surrounded by related cells. The two diagonal relations do not overlap. This means that each sub-diagonal relation constitutes a Boolean sub-algebra consisting of \(2^4\) elements, and the only shared elements, i.e. quantum entanglements, are the greatest element and the least element.

In fact, if we select the elements obtained from two different diagonal sub-relations and set them as \(\{A_1, A_8\}\) and apply the closure operator, we get

\begin{equation}
    R_*(K^*(\{A_1, A_8\}))=R_*(S_2)=S_1.
\end{equation}

This shows that it is not a fixed point. Therefore, the two obtained Boolean sub-algebras are parallelized so that they are almost independent. The Hasse diagram of the lattice obtained in this way is shown in Fig. 5.

Looking at the Hasse diagram, we can see that the two Boolean sub-algebras are divided into active and inactive phases. The Boolean sub-algebra has \(2^4\) elements, which means that four molecules are distinguished from each other, and their combination means molecular aggregation. Clustering and de-clustering defined in Token computing correspond to these active and inactive phases, so in the Hasse diagram in Fig. 5, de-clustering is represented by an arrow from the maximum cluster, which is the greatest element in the active mode, to the monomer (atom), and clustering is represented by an arrow from the monomer to the maximum cluster in the inactive mode. The Hasse diagram shows the interaction between the chemical reaction (activation and inactivation) and its execution environment (clustering and de-clustering) shown in Token computing. In other words, de-clustering proceeds in active mode, inactivation occurs due to entanglement with the least element, clustering proceeds in inactive mode, activation proceeds through entanglement with the greatest element, and oscillations occur as this process is repeated.

\begin{figure}
    \centering
    \includegraphics[width=1\linewidth]{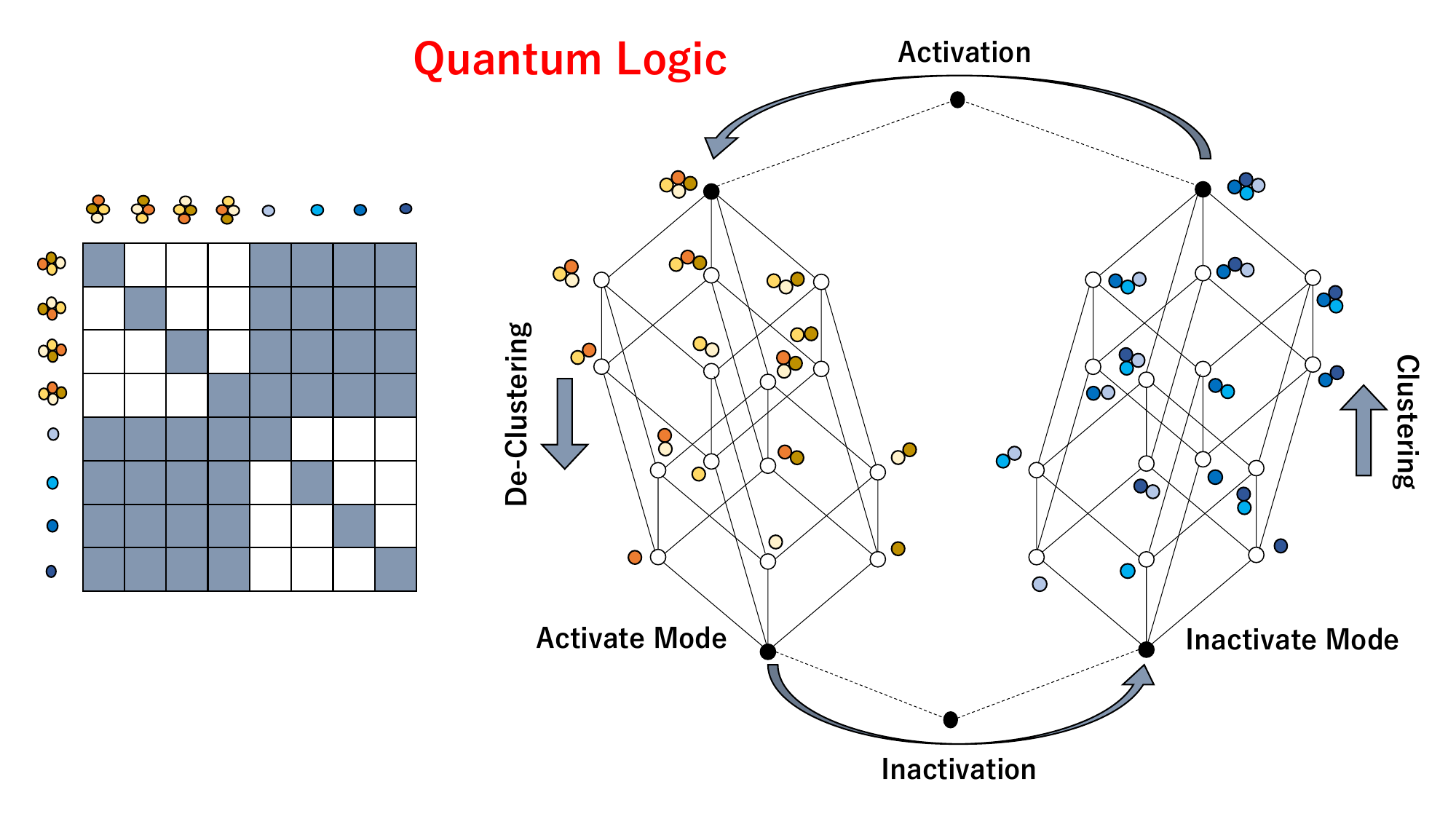}
    \caption{A lattice structure (represented as a Hasse diagram) obtained from a binary relation consisting of diagonal sub-relation and its surrounding relation cell, and a closure operation consisting of rough set approximations. De-clustering process of active molecules and clustering process of inactive molecules are illustrated by thick arrows. Activation and inactivation are illustrated by thick arrows via entanglement (shared elements). Solid lines represent order relation and dotted line represents equivalence relation in Hasse diagram}
    \label{fig:enter-label}
\end{figure}

In other words, de-clustering in the active mode and clustering in the inactive mode are set to be separated by assuming them to be different Hilbert spaces, which means that different Hilbert spaces are connected only through the situation of monomerization of all molecules and the maximum cluster composed of all molecules. A perturbation is something that introduces minute confusion and interaction into these different separated Hilbert spaces.
Also, here, we assume only four molecules and separate the active and inactive modes. However, we perform numerical simulation using 200 molecules in the next section. In other words, a chemical reaction consisting of 200 molecules is composed of a pair of Boolean sub-algebra consisting of \(2^{200}\) elements. For simplicity, the Hasse diagram shown in Fig. 5 considers a chemical reaction consisting of only four molecules.

\section{Quantum coherence in Open Computing of Chemical Reaction}\label{sec12}

\subsection{Interplay of Token and Type computing}\label{subsec1}

Ultimately, Open computing is composed of the superposition of Token computing and Type computing (Fig. 6). In Token computing, de-clustering of active molecules and clustering of inactive molecules are defined separately, and these are connected by activation in maximal clusters and inactivation in monomers, thereby implementing the mixture of computation and the computational environment. Furthermore, by introducing fluctuations—random inactivation of active molecules and random activation of inactive molecules—the invalidation of computation and the computational environment is implemented, and Open computing is achieved through the coexistence of mixture and invalidation. To achieve this implementation, fluctuations are constantly introduced from outside. Therefore, this system is essentially open.

What is the role of Type computing? At first glance, it seems like it simply approximates Token computing to find quantum logic. In other words, Token computing is a model that represents real-world physical chemical reactions, while Type computing seems merely an epistemological model that interprets those physical phenomena. However, we have stated that this is not the case; both Token computing and Type computing are phenomena and perceptions, and the emphasis on one is a matter of degree. In other words, Type computing is not an approximation by human, but in a sense an action realized by matter at a macro level, realizing a phenomenon similar with that which humans’ experience at a material level.

\begin{figure}[h]
    \centering
    \includegraphics[width=1\linewidth]{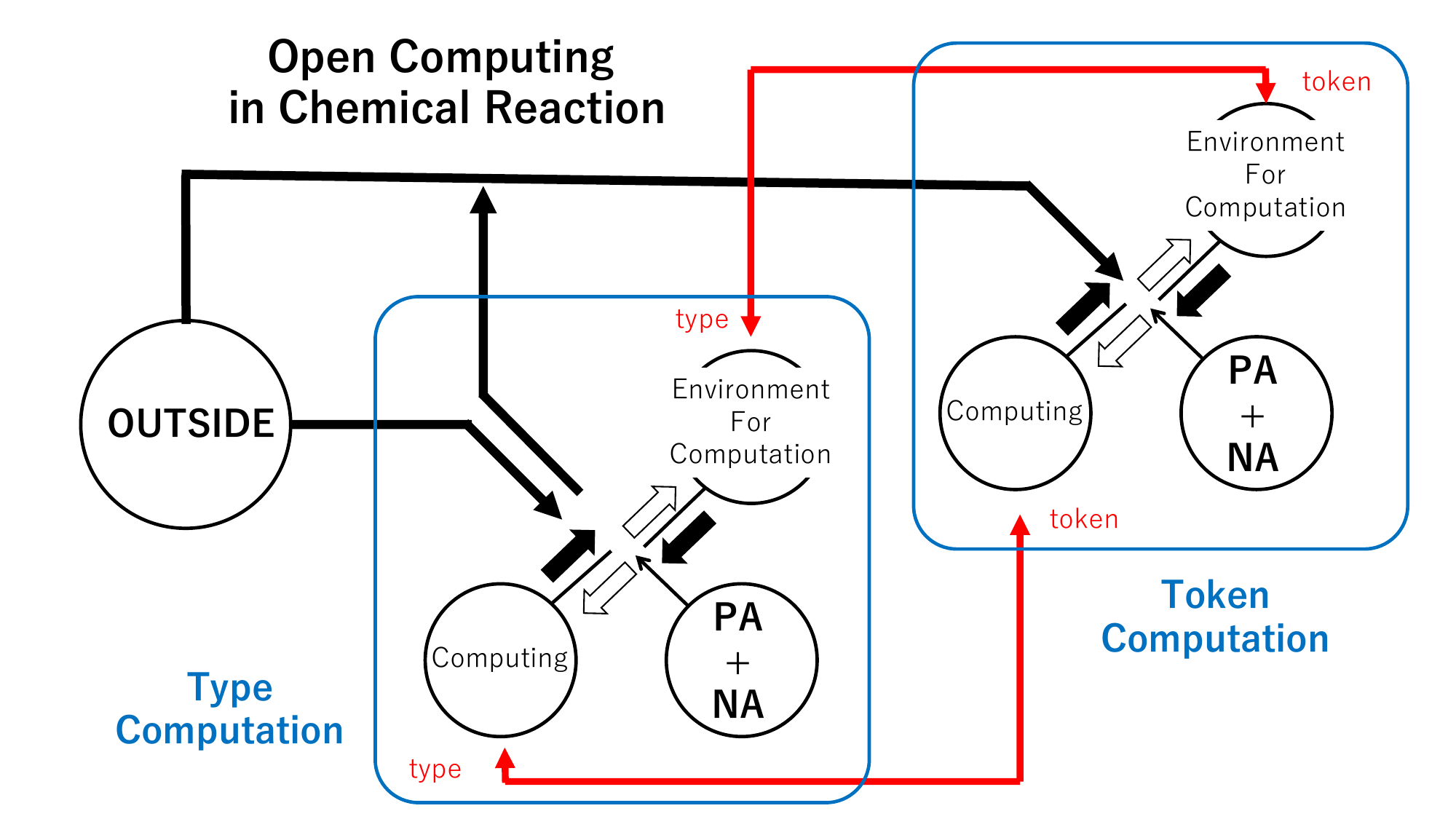}
    \caption{Schematic diagram of the interplay of Token and Type computing. PA and NA are abbreviations for ``Positive Antinomy" and ``Negative Antinomy". Since computing and its execution environment are logically different from each other, the mixture of them leads to positive antinomy. The invalidation of them leads to negative antinomy. See text for the details.}
    \label{fig:enter-label}
\end{figure}

It results in the Open computing as a model, as shown in Fig. 6, Type computing works to adjust fluctuations from outside. However, even though it adjusts fluctuations from outside, the originally tiny fluctuations that arise are implemented and introduced by disabling the computations and computational execution environment in Token computing, and Type computing amplifies the fluctuations thus introduced, affecting Token computing. In other words, the influence from Type computing that affects the fluctuations from outside to Token computing, as shown in Fig. 6, is an influence on the fluctuations that result from Token computing. In that sense, Fig. 6 illustrates the overlap of Token computing and Type computing as passive and active influences on the outside, i.e., on fluctuations.

So what exactly is Type computing, which adjusts fluctuations? Let us reconsider the lattice obtained by approximating using rough sets in Type computing. Fig. 7 illustrates its implications in the context of chemical reactions. Type computing defines a transition rule for the states of chemical species as binary relations. These transitions show how a given combination of chemical species changes, and Fig. 7 shows these with arrows that change from one colored circle to another. The process of determining necessary and sufficient conditions for the cause and effect of this reaction and collecting only stable reactions in that sense is the task of collecting fixed points with respect to a closure operator defined by the composition of upper and lower approximations. 

\begin{figure}[h]
    \centering
    \includegraphics[width=1\linewidth]{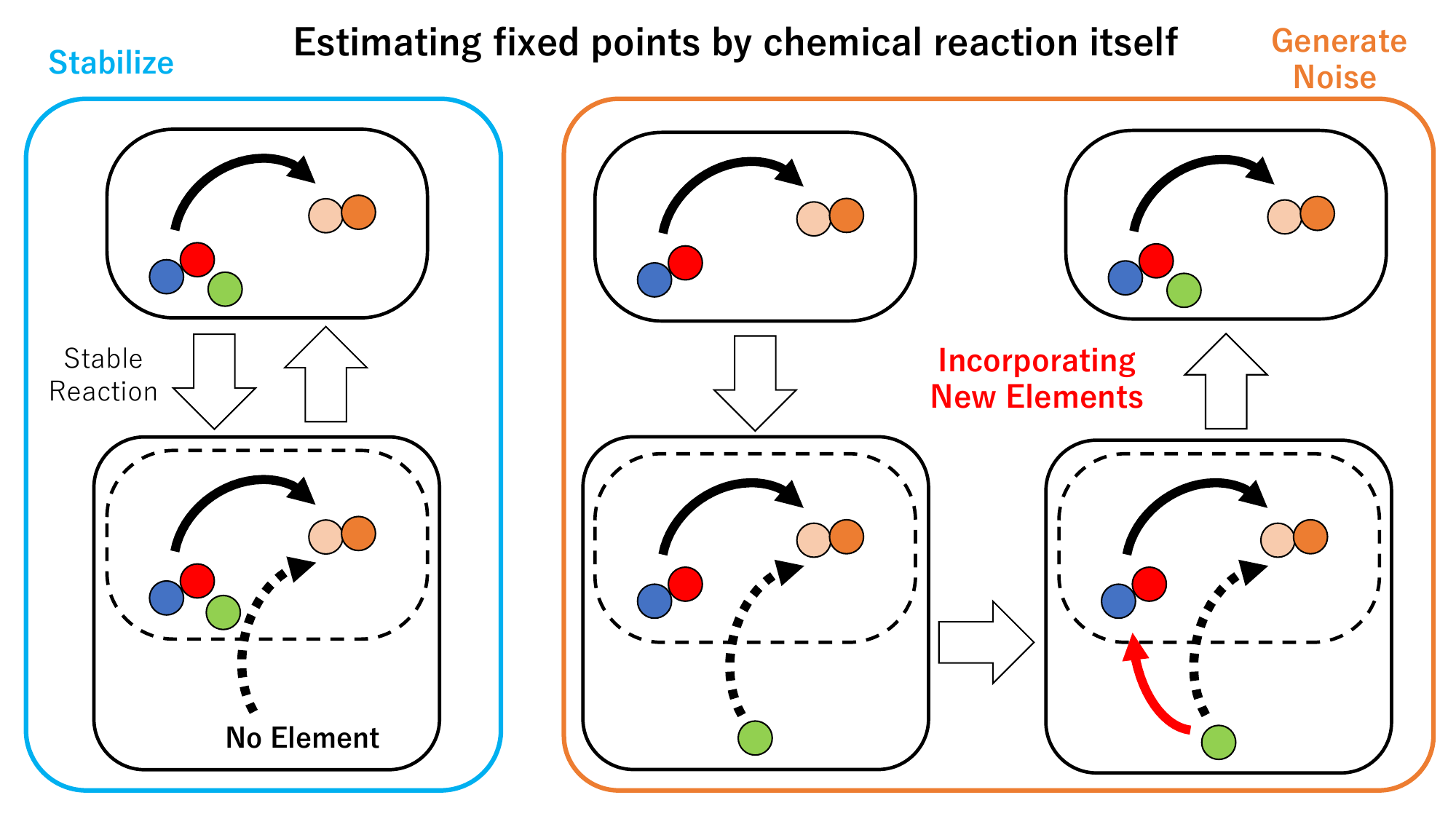}
    \caption{A diagrammatic illustration of a model that evaluates fixed points (stable chemical reactions) and non-fixed points (unstable chemical reactions) at the material level. A diagram surrounded by blue loop represents stable reaction (fixed points) and a diagram surrounded by red loop represents instable reaction (not fixed point).}
    \label{fig:enter-label}
\end{figure}

First, we determine the set of chemical species that cause the reaction. This is equivalent to determining a subset to which the closure operator can be applied. Next, we list the possible outcomes that result from this. This is equivalent to calculating an upper approximation. Then, we calculate whether there are any other causes that could lead to these possible outcomes. This is equivalent to calculating a lower approximation. If no new causal chemical species other than the initial causes are found, we can determine that the initial causes were sufficient. This means that they were fixed points with respect to the closure operator (see the blue box on the left of Fig. 7). 

On the other hand, the diagram in the red box on the right of Fig. 7 is an example that does not become a fixed point. Here, red and blue molecules (shown as circles) are used as the chemical species that cause the reaction. An upper approximation reveals that the resulting product is flesh and brown molecules. A lower approximation then checks to see if there are any other substances that cause this product. In this case, it turns out that not only the red and blue molecules, but also green molecules, can produce the same product. In other words, the red and blue molecules initially assumed were not sufficient for the chemical reaction. This means that it was not a fixed point with respect to the closure operator. In the lattice generation operation, such product candidates that are not fixed points are discarded as they do not contribute to a stable chemical reaction. In other words, the lattice obtained by rough sets is a system obtained as a collection of stable chemical reactions.

Token computing implements the phase separation of de-clustering by active molecules and clustering by inactive molecules, as well as the connection and transition of the two phases of activation and inactivation processes (which corresponds to invalidation). However, if only stable reactions are collected using Type computing, phase separation is implemented as a Boolean sub-algebra, and the activation and inactivation process is realized only through entanglement between two phases, and the transition between two phases, which corresponds to invalidation, is eliminated as an unstable factor.

\begin{figure}[h]
    \centering
    \includegraphics[width=1\linewidth]{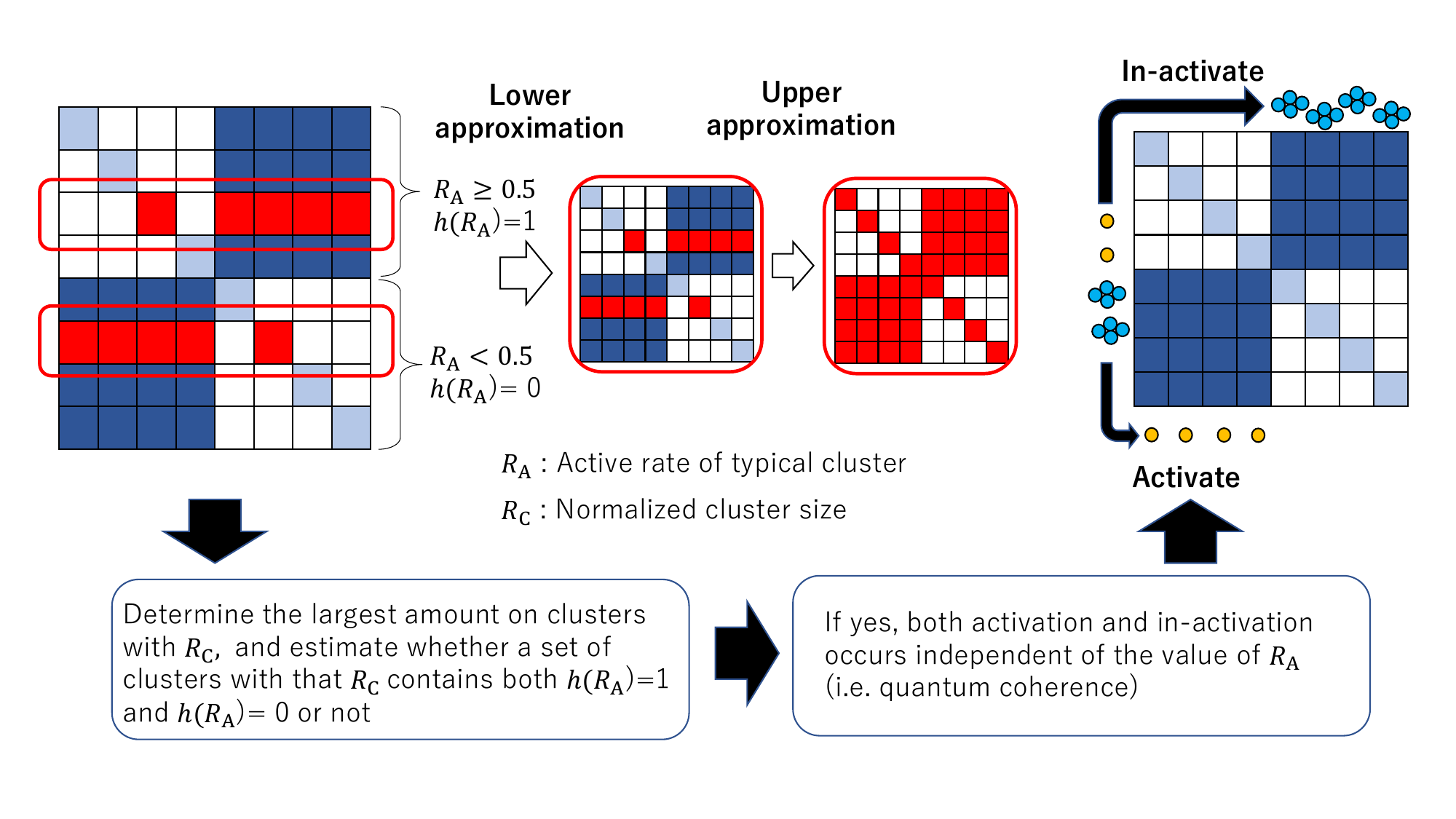}
    \caption{Explanation of binary relations (transition matrices) when a fixed point is not reached for closure operations, and the state diverges to the entire set, amplifying noise. See text for the details}
    \label{fig:enter-label}
\end{figure}

However, as mentioned above, Type computing is conceived as both a cognitive and a phenomenon. This means that the task of collecting fixed points must be implemented at the material level, not by humans. Fig 7 indeed implies that the process of collecting fixed points is realized at the material level, rather than by humans. First, the operation of collecting stable fixed points at the material level is implemented as an operation that checks whether the system is stable and, if so, eliminates noise. Second, the operation of determining whether a fixed point has not been reached is implemented not as an operation that eliminates non-fixed points, i.e., unstable reactions, but as a process of recruiting unexpected molecules or reaction causes. In the example shown in the red box in Fig. 7, this means introducing the initially unanticipated green molecule as the cause of the reaction, or recruiting it as a fluctuation.

The selection of this fixed point at the material level is implemented in the chemical reaction model proposed here as shown in Fig. 8. Note that in the binary relation of Type computing shown previously, the upper half is the phase of active molecules, and the lower half is the phase of inactive molecules. This relation is used to recruit large fluctuations. It is implemented as follows: First, the frequency distribution of all clusters is examined, and the most frequent cluster, \(C_{mode}^t\), is determined, such that

\begin{equation}
    f(C_{mode}^t)>f(k), k=1, 2, …, C_{max}^t
\end{equation}{}

where \(f(x)\) denotes the frequency of a cluster \(x\). The most frequent cluster is considered a cluster type obtained by statistical sampling. The molecules that make up that type cluster are determined to be either active clusters with a high percentage of active molecules or inactive clusters with a high percentage of inactive molecules. To do this,

\begin{equation}
    R_A = C_1^t(C_{mode}^t)/ C_0^t(C_{mode}^t)
\end{equation}

is calculated. Token computing defines the mixture and invalidation of computing (activation, inactivation) and the environment for computing (clustering, de-clustering), and acknowledges the existence of minute fluctuations. Since only fluctuations allow the mixture of active and inactive molecules, the tolerance range of this fluctuation is defined as a threshold \(\theta_A\), and then

\begin{equation}
    \theta_A \le R_A \le1-\theta_A
\end{equation}

The equation \(\theta_A \le R_A \le1-\theta_A\) determines whether a type cluster contains a mixture of active and inactive molecules.

In the binary relation (transition rule) shown in Fig. 8, which illustrates Type computing, the upper row represents active molecules and the lower row represents inactive molecules. Therefore, if \(R_A\) satisfies the above condition and a type cluster contains a mixture of active and inactive molecules, this indicates that the type cluster selected as a subset \(X\) of \(S_1\) in the binary relation on the left side of Fig. 8 is the sum of the upper and lower row subsets (the red loop on the left side of Fig. 8). Applying the closure operator to this \(X\) and calculating the upper approximation (the red loop in the second relation from the left in Fig. 8) and then the lower approximation (the red loop in the third relation from the left in Fig. 8) reveals that the subset \(X\) of \(S_1\) diverges to the entire \(S_1\).

When humans select fixed points, such type clusters are rejected because they are not fixed points and do not become lattice elements. However, when selecting a fixed point at the material level, the process of divergence from \(X\) to the entire \(S_1\) becomes the recruitment of fluctuations in the unstable reaction described in Fig. 7. In other words, if \(\theta_A \le R_A \le 1-\theta_A\) holds, the regions of the relational cells, \(\{A_1, A_2, A_3, A_4\}\times \{ a_5, a_6, a_7, a_8\}\) and \(\{A_5, A_6, A_7, A_8\}\times \{ a_1, a_2, a_3, a_4\}\), will also be active simultaneously with the diagonal sub-relations. That is,

\begin{equation}
    m_1^{(t+1)}(k)=h(R_A) 
\end{equation}

under \(P_{coh}\), where \(h:[0.0, 1.0] \rightarrow \{0, 1\}\) is defined by: for \(x\in[0.0, 1.0]\), \(h[x]=1\) if \(x<0.5\); \(h[x]=0\) otherwise, and \(0.0\le P_{coh} \le 1.0\). This transition probability reflects the state transition of the region containing the relational cells. If \(R_A \ge0.5\), an active molecule becomes an inactive molecule, and if \(R_A<0.5\), an inactive molecule becomes an active molecule. This transition is shown in the right-hand relation (transition matrix) in Fig. 8.

The choice of a fixed point at the material level can amplify fluctuations in unstable reactions. This is the true significance of open computing, the interplay between Token computing and Type computing.

\subsection{Critical Phenomena and Quantum-like coherence
}\label{subsec2}

The interplay between Token computing and Type computing amplifies the small fluctuations inherent in Token computing, resulting in a phenomenon that could be described as quantum coherence. Let us examine this in turn.

\begin{figure}[h]
    \centering
    \includegraphics[width=1\linewidth]{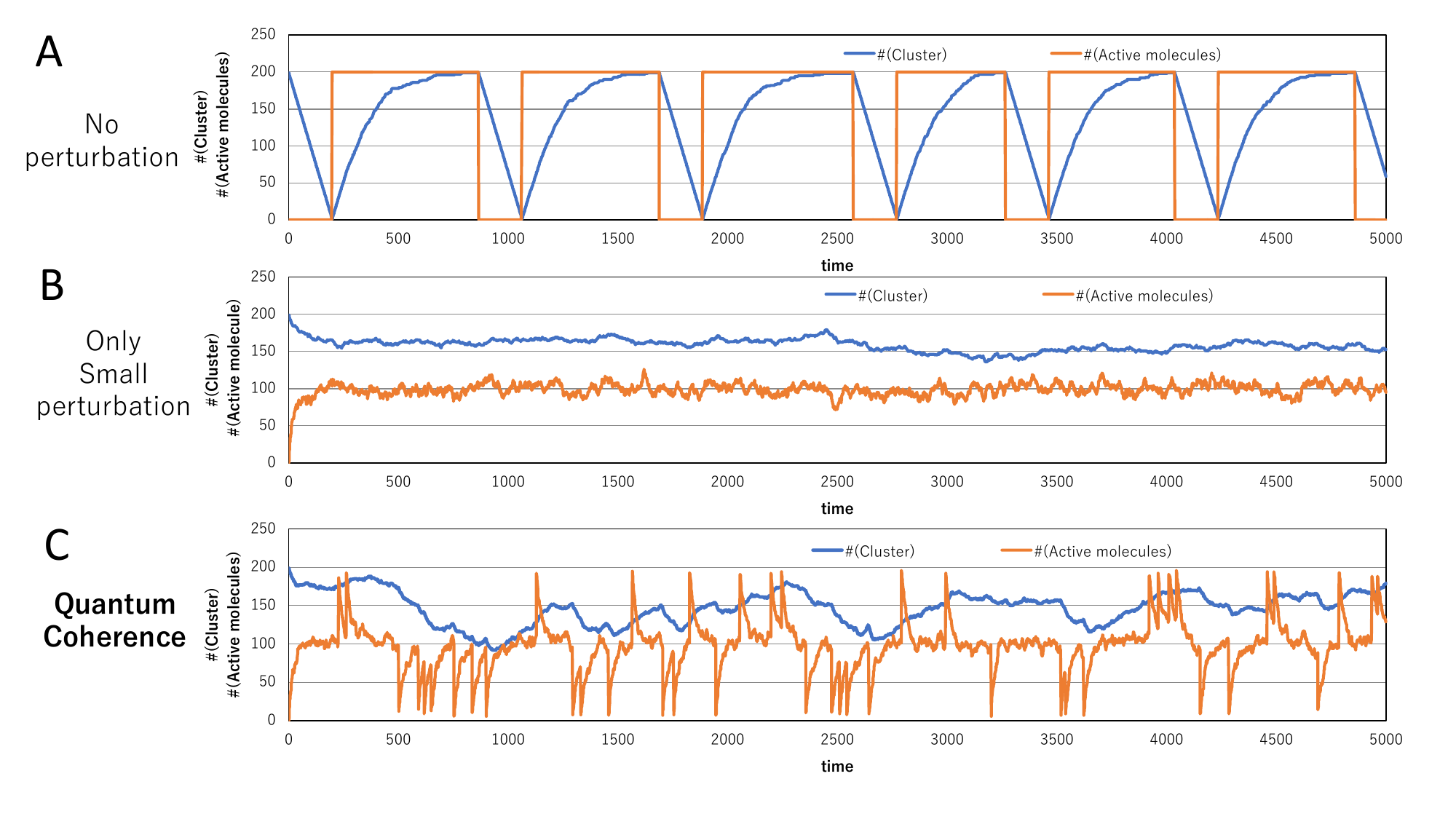}
    \caption{The number of clusters (blue curves) and the number of active molecules (orange curves) in Open Computing of chemical reactions are plotted against time.
A. Token computing without noise.
B. Token computing under noise with \(P_{noise}=0.05\)
C. The interplay between Type and Token computing results in drastic transitions between inactive and active molecular states, producing intermittent spike-like wave patterns.}
    \label{fig:placeholder}
\end{figure}

The three graphs in Fig. 9 show the results of an open computing simulation of a chemical reaction as defined in this paper. The blue curve plots the number of clusters against time, and the orange curve plots the number of active molecules against time. Fig 9A shows the results of a time evolution calculation using a model that implements only a mixture of chemical reactions (activation and inactivation) and their execution environment (clustering and de-clustering) without the fluctuations that Token computing typically imposes. Fig 9B shows a model in which fluctuations are added to the same situation as A with a probability of 0.05, but interplay between Token and Type computing is prohibited. Fig 9C shows a model that similarly applies fluctuations with a probability of 0.05, samples clusters using this, and determines whether active and inactive molecules are mixed, thereby amplifying the fluctuations in the transition matrix (which becomes a binary relationship through binarization). This model implements the interplay of Token and Type computing, and is the original Open Computing model.
However, in all cases, the number of molecules is 200. The parameters are also set to essentially the same values. The cluster activation rate, which is the condition for clustering, is \(\theta_{c}\) = 0.5 (below this, polymerization occurs), and the cluster activation rate, which is the condition for de-clustering, is also \(\theta_{dec}\) = 0.5 (above this, decomposition occurs). In Fig. 9A, where there is no fluctuation, \(P_{noise}\) = 0. In Figs. 9B and C, where fluctuations are introduced, \(P_{noise}\) = 0.05. Furthermore, the parameter that realizes the interplay between Type computing and Token Computing in Fig. 9C is \(\theta_A\) = 0.3. When not at a fixed point, noise is enhanced, and the probability of deactivation or activation is \(P_{coh}\) = 0.95.

In the absence of fluctuations, chemical reactions and the reactions connected to realize the mixture of their execution environment proceed alternately. Here, the initial state is set to 200 inactive molecule monomers, resulting in 200 clusters and 0 active molecules. At each step, all clusters are assigned a number, and two are randomly selected from them for polymerization. Therefore, polymerization occurs once per time step, resulting in a monotonically decreasing cluster count. When all monomers aggregate to a single cluster, all molecules become active molecules. As long as the molecules remain active, the large clusters are de-clustered. De-clustering is performed by randomly selecting one monomer from the total and splitting it into two. If a divisible cluster is selected, the number of clusters increases through splitting. However, if a monomer is selected, the number of clusters does not increase because it cannot be split. Therefore, as de-clustering progresses and the number of monomers increases, the rate of cluster growth slows. De-clustering continues until de-clustering is no longer possible, and all clusters become monomers. At that point, all molecules become inactive monomers, returning to the initial state. The same process then repeats. When there are no fluctuations, this type of oscillation repeats.

Fig 9B shows the time evolution of a chemical reaction when fluctuations are introduced into this model with a probability of \(P_{noise}\) = 0.05. With this probability, active molecules become inactive, and inactive molecules become active. This change inhibits clustering and de-clustering. Clustering progresses when the cluster activation rate is below 0.5. Without fluctuations, clustering occurs in a state of inactive molecules, so this condition is met. Similarly, de-clustering occurs when the cluster activation rate is above 0.5. Without fluctuations, all molecules are active, so this condition is also met.

\begin{figure}[h]
    \centering
    \includegraphics[width=1\linewidth]{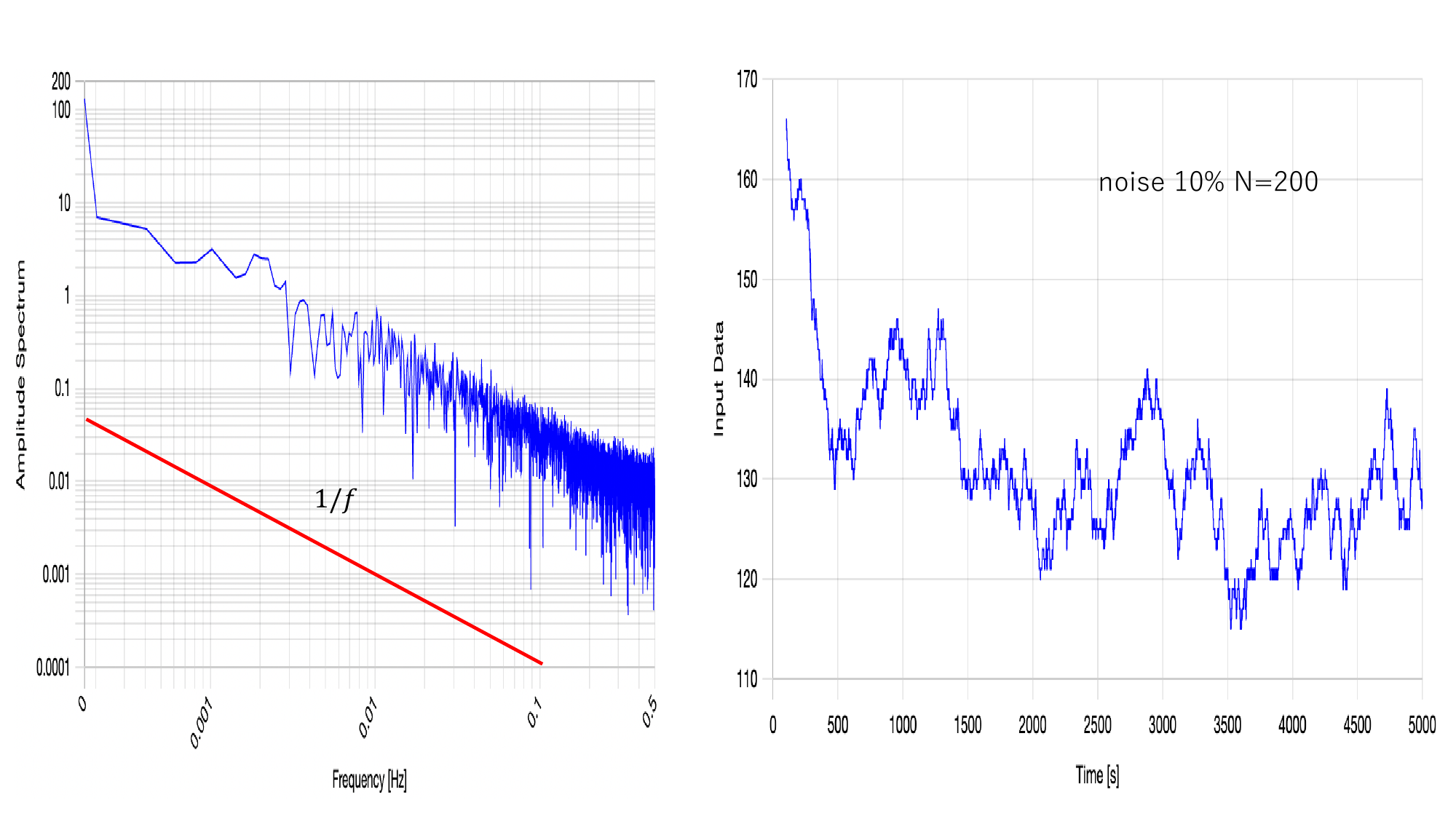}
    \caption{Power spectrum of the time series of the number of molecules in Token computing, plotted against frequency on a log-log scale (left). The corresponding time series of molecular count is shown on the right.}
    \label{fig:placeholder}
\end{figure}

However, when fluctuations are introduced, the clustering and de-clustering conditions are immediately overturned by the fluctuations. Once clustering progresses to a certain extent, it is overturned. Conversely, even if de-clustering progresses, it will stop at a certain point. Because the point at which it stops is determined by the nature of the fluctuations, the number of clusters and active molecules is not random. In fact, when we Fourier transform this time series and plot the power spectrum logarithmically against frequency, we obtain 1/f noise with a slope of -1 in the strict sense (Fig. 10). In other words, Token computing, which inherently contains fluctuations, autonomously realizes critical behavior between chaos and order in the sense of randomness, demonstrating self-organized criticality \cite{Bak-Tang1989, Bak-Sneppen1993, Langton1990, Kauffman-Johnsen1991}.

Fig 9C shows the time evolution when the introduced fluctuations are not fixed points in Type computing, i.e., when fluctuations are amplified by unstable chemical reactions. Here, spikes occur intermittently. The timing of their occurrence depends on the change in the number of clusters: upward spikes occur when the number of clusters increases rapidly, and downward spikes occur when the number of clusters decreases rapidly. When clusters increase rapidly, the clustering condition is maintained for a short period of time, and while inactive and active molecules coexist, the dominance of inactive molecules is maintained. In other words, because fluctuations are amplified in the lower inactive molecule region of Type computing, molecules rapidly become activated, causing a sudden increase in active molecules. However, the sudden increase in the number of active molecules amplifies fluctuations in the active molecule region, which in turn rapidly accelerates deactivation. This sudden increase in the number of active molecules immediately reverses to a decrease, resulting in a spike. Downward spikes are generated by the exact opposite process.

The generation of these spikes can be described as quantum-like coherence. In the absence of fluctuations, the active phase, which realizes de-clustering, and the inactive phase, which realizes clustering, are separated. The two phases, where completely different conditions hold, are understood as two Boolean sublattices in the quantum lattice. They correspond to two Hilbert spaces that normally cannot interact. Here, the process of amplifying fluctuations through unstable reactions and recruiting large fluctuations represents a transition from the active phase to the inactive phase, and vice versa. In quantum mechanical terms, this represents interference between different Hilbert spaces. As a result, a strong correlation between the two Hilbert spaces occurs, i.e., either a sudden activation or inactivation occurs in both, resulting in spike waves. In this sense, a clear coherence is observed here, and we can use the term quantum-like in the Khrenikov sense to call it quantum-like coherence~\cite{khrennikov2015quantum,khrennikov2008quantum}.

These spikes suggest that spikes can be generated and signaled in specific chemical reaction systems other than neurons. The most likely candidate is the spike generated by proteinoid microspheres~\cite{fox1959production,fox1960thermal,przybylski1984excitable,przybylski1986electrical,mougkogiannis2023low}. Proteinoid microspheres form through the reaction of amino acids and bases, undergoing repeated polymerization and depolymerization~\cite{matsuno1984electrical}. The proteinoid microspheres generate action-potential like spikes. Also, when measured with macro-electrodes EEG like slow waves are evidenced. Patterns of electrical activity of proteinoid ensembles can be moduled using optical, electrical and chemical stimulation. 
Research is underway to explore how this can be used for computational signal transduction~\cite{mougkogiannis2023logical}. It is highly likely that quantum-like coherence, as described in this paper, is involved in the formation of these spikes.

\begin{figure}[h]
    \centering
    \includegraphics[width=1\linewidth]{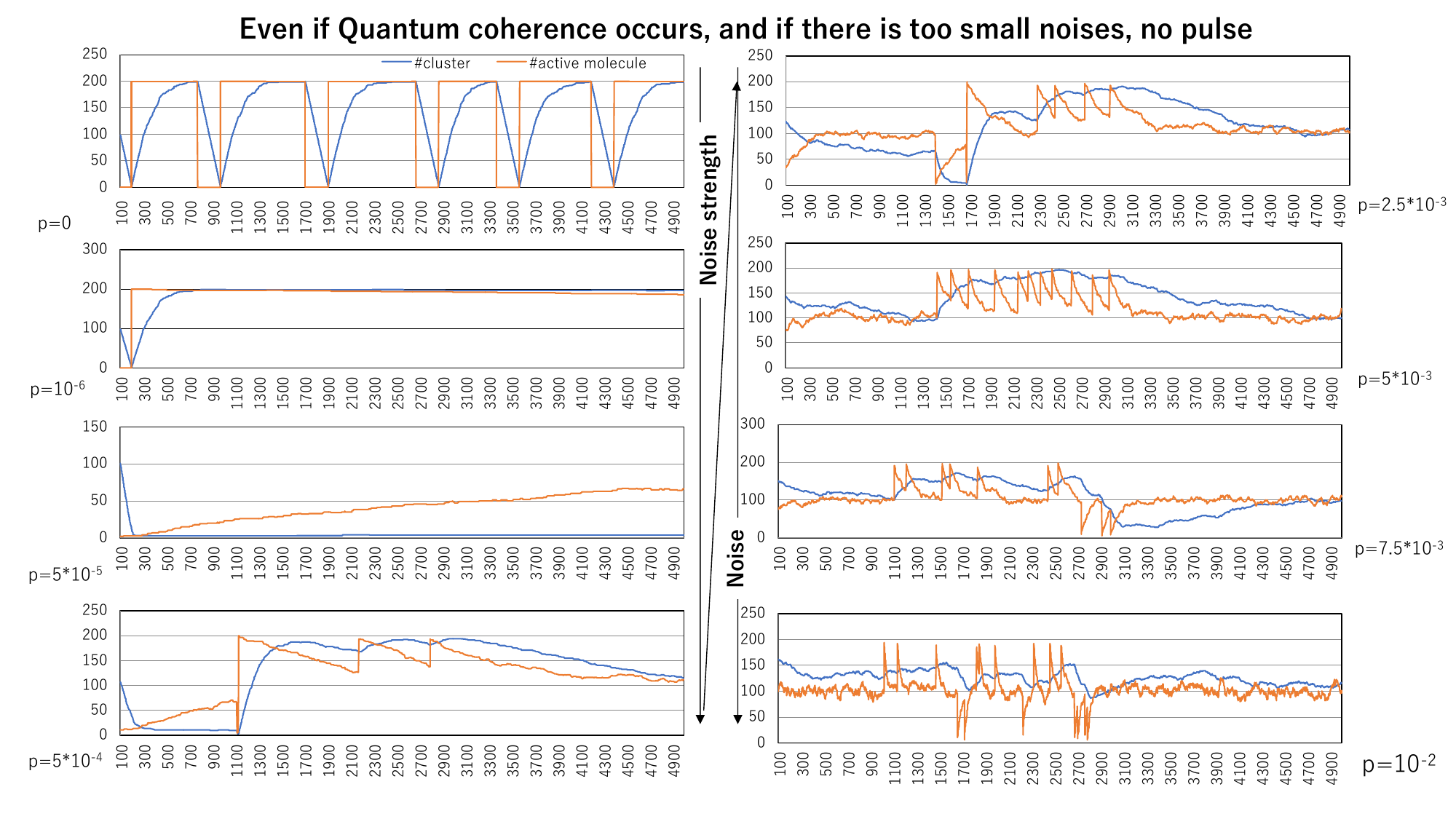}
    \caption{Effects of the Interplay Between Type and Token Computing Under Varying Noise Strengths.
The temporal dynamics of the number of clusters (blue) and active molecules (orange) are modulated by the intensity of noise. Time series data corresponding to each noise level are shown, with the respective noise strength indicated below each plot.}
    \label{fig:placeholder}
\end{figure}

Fig. 11 evaluates how the magnitude of fluctuations affects spike formation under conditions in which interplay between Type computing and Token computing is active. In the absence of fluctuations, interplay is possible, but no mixture of active and inactive molecules is observed in any cluster, resulting in the same oscillations in the number of clusters and active molecules as in Fig. 9A (Fig. 10 top left). From here, the noise intensity increases in the direction of the arrows. When the noise intensity is \(P_{noise}=10^{-6}\), the number of clusters does not decrease even after it increases, and remains high along with the number of active molecules. This is because even if the fluctuation is small, the inactivation of even one molecule does not create the conditions for simultaneous inactivation of all monomers, preventing the transition to de-clustering. When the noise intensity further increases to \(P_{noise}=5 \times 10^{-4}\), in very rare cases, clustering is continuously maintained and a sudden increase is realized. In this case, the conditions for the formation of a spike wave are met, and the inactive molecules are simultaneously activated, followed by the simultaneous inactivation of the active molecules, resulting in a spike wave. However, because successive clustering and de-clustering events are extremely rare, the frequency of spike waves is extremely low. Furthermore, even if a sudden activation of inactive molecules occurs, it is even more difficult for a sudden inactivation of active molecules to occur afterwards. Therefore, in most cases, the waves generated do not become spike waves, but rather sawtooth waves. This tendency does not change even when the noise intensity increases further, from \(P_{noise}=5 \times 10^{-3}\) to \(7.5 \times 10^{-3}\), and it can be seen that sawtooth waves are continuous, rather than spike waves. As the noise intensity increases further, a series of clustering and de-clustering that meet the conditions is realized, and spike waves, rather than saw-tooth waves, become dominant.

\begin{figure}[h]
    \centering
    \includegraphics[width=1\linewidth]{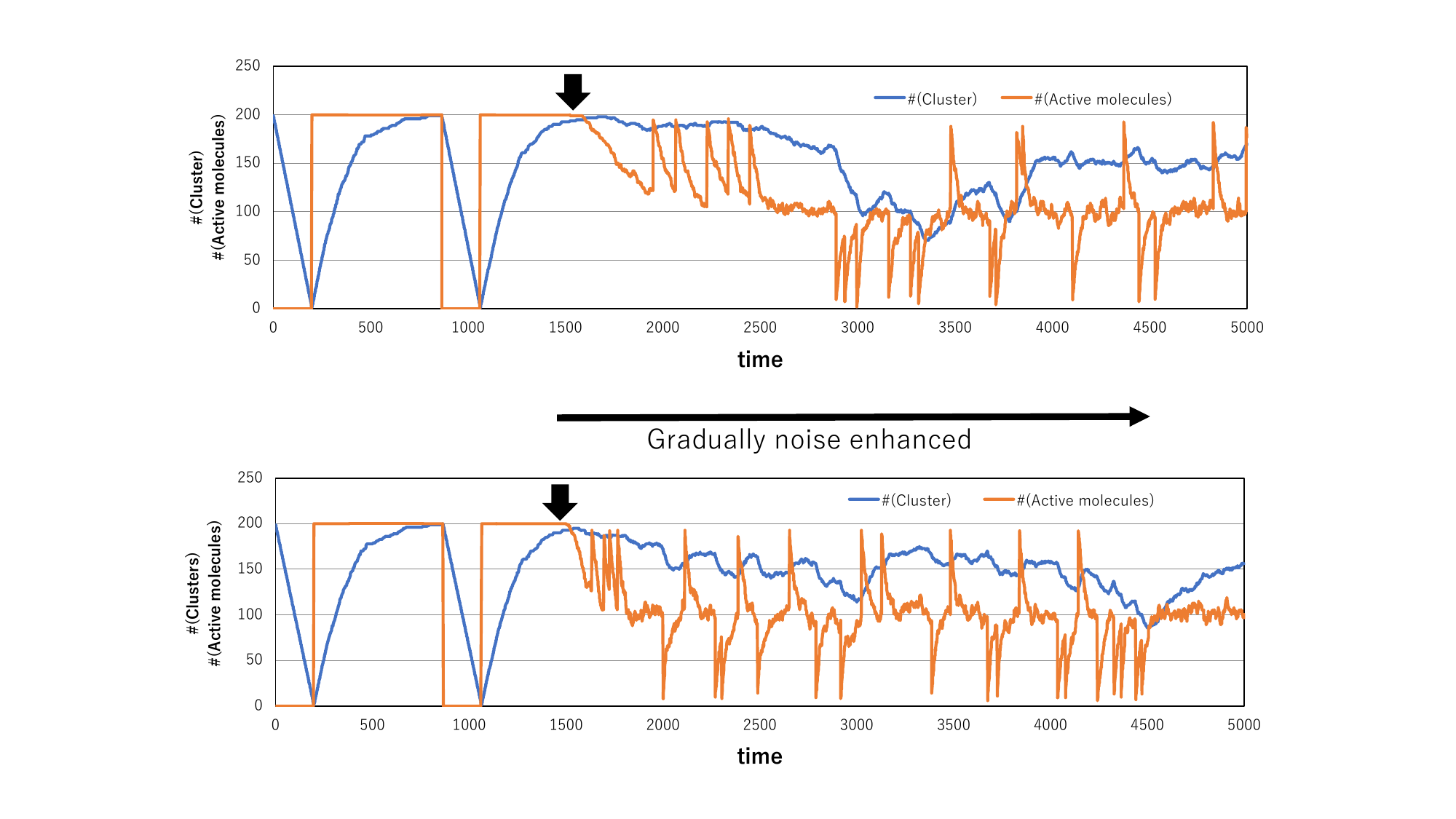}
    \caption{Time series of clusters and active molecules under gradually increasing noise.
Time series of the number of clusters (blue) and active molecules (orange) are plotted as a function of time, illustrating the effects of gradually increasing noise. The onset of noise injection is indicated by the bold arrow.}
    \label{fig:placeholder}
\end{figure}

Fig 12 shows a case in which the noise intensity gradually increases over time. As with Figs. 9 and 11, this Fig shows the time evolution of the number of clusters and the number of active molecules. However, noise is injected at the point indicated by the thick arrow, and the noise intensity gradually increases thereafter. Because the interplay between Type computing and Token Computing is implemented here, once the fluctuations reach a certain magnitude, the critical behavior seen in Fig. 9B is not observed. Instead, a small sawtooth wave immediately appears, and as the noise intensity increases, a spike wave is immediately generated. The increase in noise intensity is more gradual in the chemical reaction in the upper panel than in the lower panel, so a series of sawtooth waves appears before the transition to a spike wave. This shows how changes in noise intensity affect the appearance and frequency of spike waves.

As described above, open computing, implemented as an interplay between Type computing and Token computing, allows critical phenomena to occur latent and further amplifies their noise, thereby realizing the interaction of Hilbert spaces (here, Boolean sublattices) that were originally separated, and realizing what might be called quantum-like coherence. This results in spike waves.

Such spike waves typically require an autocatalytic reaction, such as a rapid transition to or promotion of a chemical reaction. This is often achieved by enzymes.

The quantum-like coherence considered here is a pre-enzyme reaction. If real chemical reactions are realized as open computing, there is a possibility that a quantum-theoretical structure lies behind them, which would, in principle, bring about quantum-like coherence. However, this requires the absence of fluctuations. However, if an agent (i.e., a chemical species) capable of rapid activation or deactivation were to emerge, it would take over the role of quantum-like coherence and produce spike waves even in the absence of the necessary fluctuations. Conversely, quantum-like coherence is thought to be a preparation for the arrival of new enzymes. Innovation, such as the origin of enzymes, requires preparation to accept innovation, and it is thought that this is achieved by a latent quantum-like mechanism.

\section{Discussion}\label{sec13}

One author calls the method of understanding that relates binary oppositions "artificial intelligence." He proposes a new intelligence that accepts both binary oppositions, establishing a positive antinomy, and then bleaches their meaning and intensity to create a negative antinomy. This new intelligence, which accesses the external world of the two originally conceived through the coexistence of positive and negative antinomy, is called "natural born intelligence" \cite{gunji5209533natural, gunji2022psychological}. Here, closed computing and open computing, which specialize in computation, correspond to artificial intelligence and natural born intelligence, respectively. Open computing is a computational system that accesses the external world.

In particular, open computing, conceived as an interplay between Type computing and Token computing, realizes interactions between different Boolean sub-algebras in quantum logic (different Hilbert spaces in quantum mechanics), achieving quantum-like coherence. This suggests the origins of signal transduction in chemical reactions and the origins of enzymes, opening up new research possibilities.

\begin{figure}[h]
    \centering
    \includegraphics[width=1\linewidth]{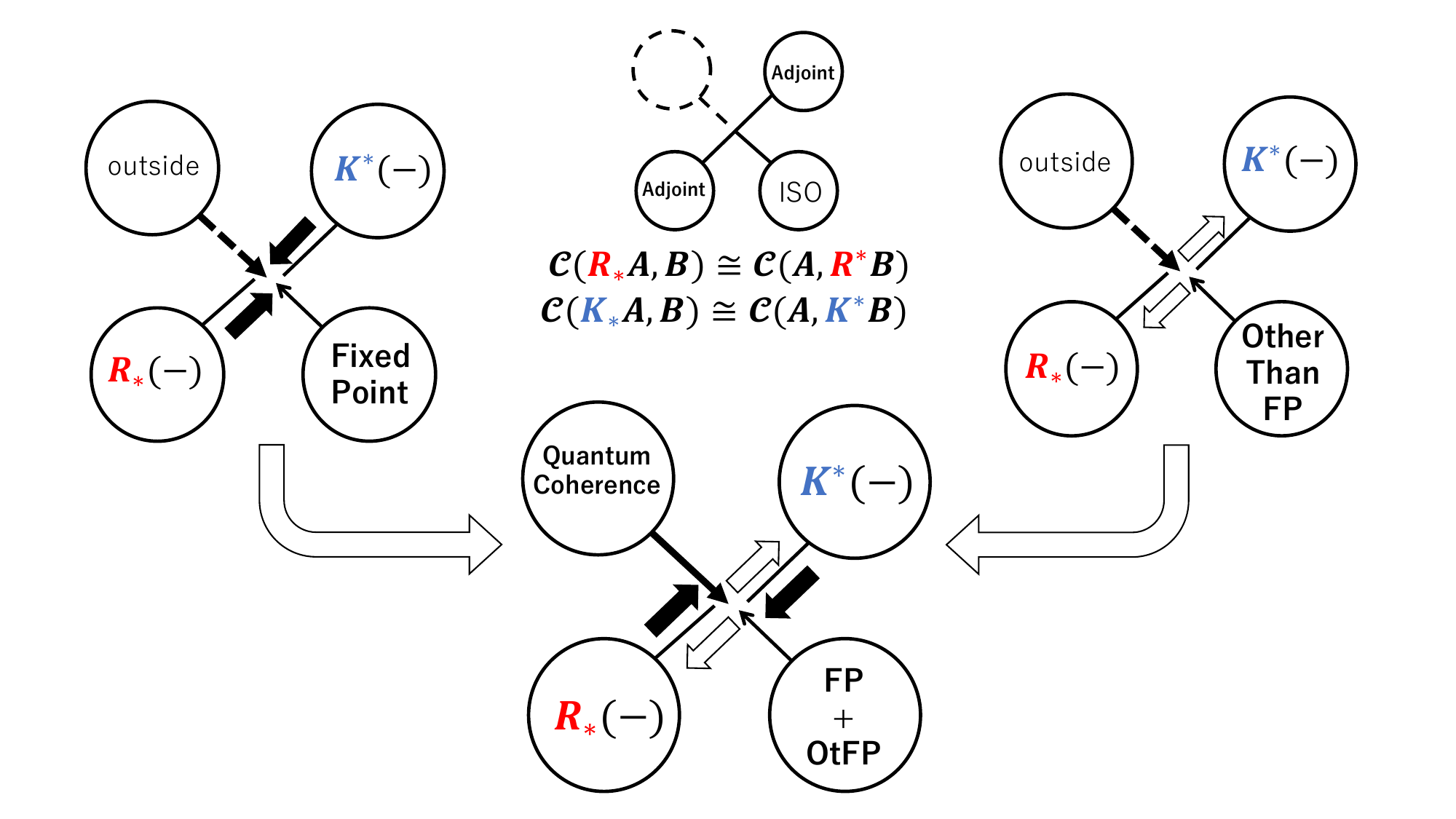}
    \caption{Schematic representation of Open computing from a Category-Theoretic perspective.
While the upper and lower approximations in rough set theory give rise to an isomorphism known as an adjunction (top center), this structure corresponds to closed computing. In contrast, Open computing emerges either through the selection of fixed points (left) or through the consideration of subsets that do not yield fixed points (right). These mechanisms collectively give rise to a structure characterized by quantum-like coherence.}
    \label{fig:placeholder}
\end{figure}

In this section, we examine open computing's quantum-like coherence from the perspective of category theory \cite{awodey2010category}. This is because category theory is, in a sense, the ultimate system of closed computing. Closed computing is calculation that somehow relates two heterogeneous entities that may seem like binary oppositions. The simplest such relation is equality. Category theory is a system that equates two entities that are heterogeneous in the usual sense, but cannot be connected by equality, through extensions of equality such as isomorphism and category equivalence. The most prominent structure is adjunction, an isomorphism obtained by adjoint functor.

Adjunction is also observed in the upper and lower approximations of rough sets discussed in this paper. An ordered set \(\bold P\) is a category in which objects are set elements (in this case, the elements themselves are sets) and arrows are order relations (in this case, containment relations). Identity is defined by the reflective law of the order relation, and composition of arrows is defined by the transitive law. A functor is a transformation from category to category that preserves identity and composition of arrows. The upper and lower approximations of the equivalence relation \(R\) on \(\bold P\) are realized by functions as follows:

\begin{equation}
\bold P(R_*(X), Y) \cong \bold P(X, R^*(Y))
\end{equation}

This is precisely the picture of closed computing, as shown in the top center of Fig. 13. So how can the interplay between Type computing and Token computing presented in this paper be represented? It can be represented precisely by the structure of Open computing.

First, defining a closure operation using two different binary relations \(R\) and \(K\) and taking its fixed point is nothing more than an operation that treats two heterogeneous entities, which cannot be connected by equality or isomorphism, as the same. This corresponds to the operation in Open computing that takes a mixture of two heterogeneous entities, the computation and the execution environment. In fact, a lattice in Type computing is obtained by choosing a fixed point between the two entities \(R_*\) and \(K^*\). This corresponds to the diagram in the top left of Fig. 13.

Second, the operation of choosing a non-fixed point between \(R_*\) and \(K^*\) expands the initially adopted subset by not being a fixed point, corresponding to recruiting noise. This corresponds to invalidating the heterogeneous mixture of \(R_*\) and \(K^*\), i.e., the operation corresponding to the invalidation of the computation and its execution environment, and corresponds to the diagram in the upper right of Fig. 13.

The coexistence of the operation of choosing a fixed point and the operation of recruiting large fluctuations from a non-fixed point can be considered an implementation of open computing in the sense of the lower right of Fig. 13. We can implement open computing not by using a model that fits within category theory, but by using category theory as a subsystem of the system. In this case, we have seen an open system implemented not as a system that fits within quantum logic, but as a system that interacts with the outside world using quantum logic as a subsystem. For the first time, it became possible to understand the realization of quantum-like coherence.

\section{Conclusion}\label{sec14}

The interaction between chemical reactions (activation and inactivation) and their environments (clustering and de-clustering) is modeled through both Token and type computation. This interplay gives rise to chemical processes that underpin quantum logic, specifically the structure of a non-distributive orthomodular lattice.
From the perspective of quantum logic, the de-clustering phase (active mode) and clustering phase (inactive mode) form distinct sub-lattices. These sub-lattices share common elements representing entanglement.
Even small perturbations can trigger a drastic mixture of active and inactive modes, potentially generating coherent phases that manifest as spike trains. Physical phenomena are thus co-created through the interaction between the observer (human) and nature. However, this interplay does not imply that physical phenomena are mere approximations by humans. Rather, the interplay of nature and humans is mediated through Token and Type computing: Token computing is inherently physical, while Type computing is inherently formal.
When chemical reactions are described solely by Type computing, the result is a collection of stable reactions approximated by quantum logic. However, the dynamic interplay between Token and Type computing enables the system to capture the role of unstable reactions, which amplify perturbations and induce quantum coherence.
In this context, we refer to the "quantum-like structure" not merely in terms of quantum logic, but also in reference to its dynamic properties. From our perspective, quantum-like structures are not rooted in microscopic physical systems. Instead, they emerge from the interplay of Token and Type computing at the macroscopic level. Such structures serve as interfaces between nature and humans—not only projecting information from nature but also incorporating perturbations from it. In this sense, quantum-like structures are intrinsically open systems.
To summarize our proposal:\\
1.	Open Computing of chemical reactions is realized through the inseparability of computation and its environment.\\
2.	This is achieved by negating interactions based on separation, allowing the system to behave as if its quantum logical approximation mediates with the external world.\\
3.	Similar to how humans collect fixed points through closure operations, chemical reactions also stabilize by accumulating fixed points. Unstable reactions are not discarded; rather, they drive transformations by mixing different sub-lattices (or Hilbert spaces), giving rise to spike waves.\\
4.	Consequently, spike waves can be understood as manifestations of quantum coherence.

\backmatter

\bmhead{Supplementary information}

not applicable.

\section*{Declarations}

\begin{itemize}
\item Funding: {Not applicable}
\item Conflict of interest/Competing interests: {The authors declare no conflicts of interest.}
\item Ethics approval and consent to participate:  {Not applicable}
\item Consent for publication
\item Data availability:  {Not applicable}
\item Materials availability:  {Not applicable}
\item Code availability:  {Not applicable}
\item Clinical Trial Number: {Not applicable}
\item Author contribution: {YPG produced grand design of the model, and mainly wrote the manuscript. AA, AK and PM read and wrote the manuscript.}
\end{itemize}

\noindent
If any of the sections are not relevant to your manuscript, please include the heading and write `Not applicable' for that section.

\end{document}